\documentclass[runningheads]{llncs}
\usepackage{graphicx}

\usepackage{tikz}
\usepackage{comment}
\usepackage{amsmath,amssymb} 
\usepackage{color}
\usepackage{booktabs}
\usepackage{numprint}
\usepackage{bm}
\usepackage{xcolor}
\usepackage[accsupp]{axessibility}  

\usepackage[width=122mm,left=12mm,paperwidth=146mm,height=193mm,top=12mm,paperheight=217mm]{geometry}
\usepackage[pagebackref,breaklinks,colorlinks]{hyperref}
\usepackage{multirow}
\usepackage{cite}
\usepackage[symbol]{footmisc}

\usepackage{xcolor,colortbl}

\newcommand{\colorfirst}{255, 153, 153}
\newcommand{\colorsecond}{255, 204, 153}

\newcommand{\myparagraph}[1]{\noindent\textbf{#1}}

\usepackage{ruler}
\usepackage[width=122mm,left=12mm,paperwidth=146mm,height=193mm,top=12mm,paperheight=217mm]{geometry}

\definecolor{DeltaColor}{rgb}{0.039,0.73,0.71}
\definecolor{SetaColor}{rgb}{0.867, 0.0235, 0.376}
\definecolor{SigmaColor}{rgb}{0.98,0.45,0.0}
\definecolor{HaoColor}{rgb}{0.8,0,0}
\definecolor{AlphaColor}{rgb}{0,0,0.8}
\definecolor{BetaColor}{rgb}{0.8,0,0.8}
\definecolor{GammaColor}{rgb}{0.5,0,0.7}
\definecolor{EpsilonColor}{rgb}{0.353,0.725,0.906}
\definecolor{TauColor}{rgb}{0.423,0.235,0.192}

\begin{document}
\pagestyle{headings}
\mainmatter

\title{KeypointNeRF: \\Generalizing Image-based Volumetric Avatars using Relative Spatial Encoding of Keypoints}

\titlerunning{KeypointNeRF}

\author{
Marko Mihajlovic$^{1,2}$\thanks{The work was primarily done during an internship at Meta.}, Aayush Bansal$^{2}$, Michael Zollhoefer$^{2}$, Siyu Tang$^1$, Shunsuke Saito$^{2}$\\
$^1$ETH Z\"{u}rich \ $^2$Reality Labs Research
}
\institute{
{\href[pdfnewwindow=true]{https://markomih.github.io/KeypointNeRF}{\nolinkurl{markomih.github.io/KeypointNeRF}}}
}
\authorrunning{M. Mihajlovic et al.}
\maketitle

\begin{center}
    \centering
    \captionsetup{type=figure}
    \includegraphics[width=\textwidth]{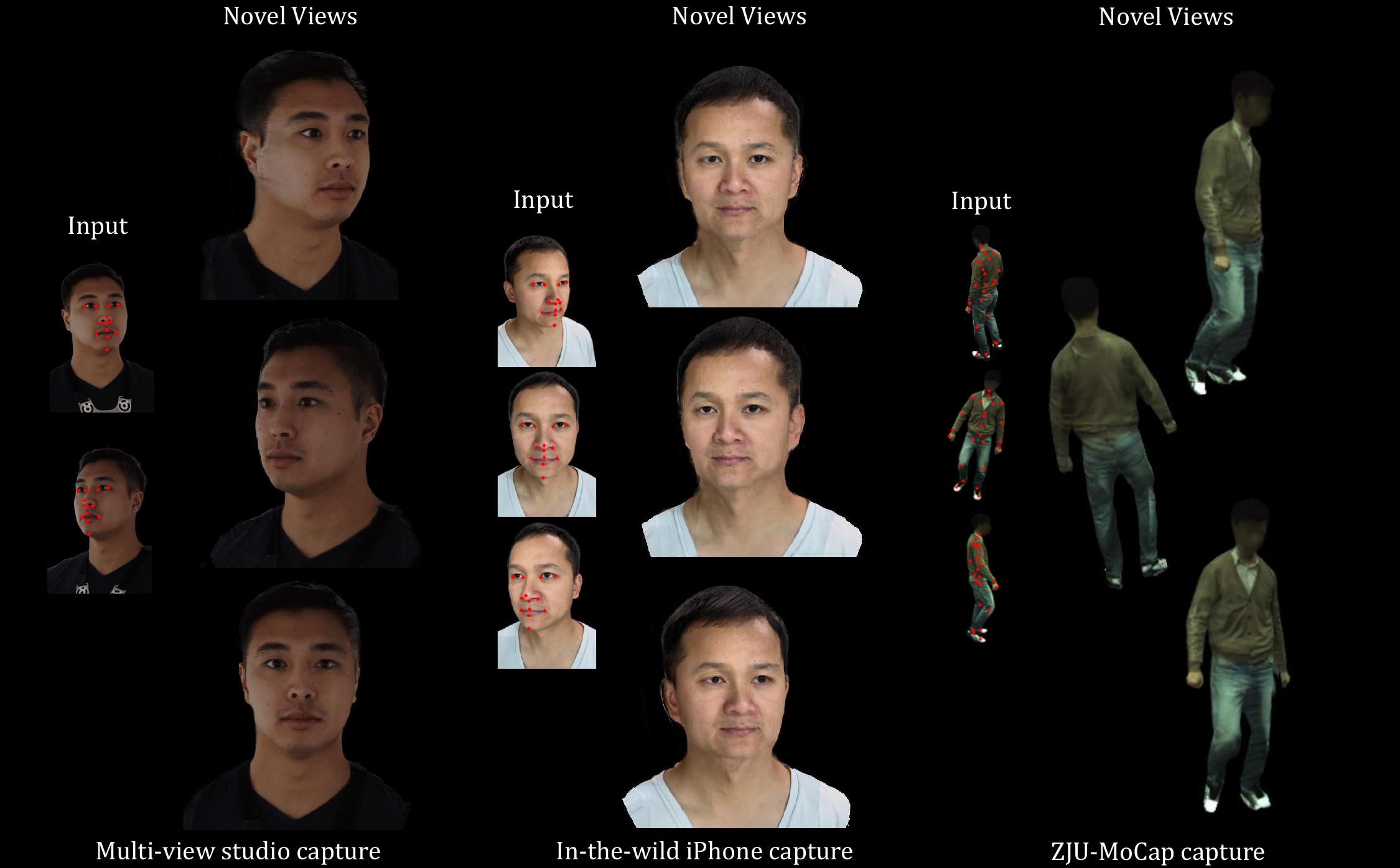}
    \captionof{figure}{\textbf{(a)} Our approach allows us to synthesize high-fidelity volumetric humans from two or three views of unseen subjects. \textbf{(b)} Model learned on studio captures can be used without modification on in-the-wild iPhone captures; and \textbf{(c)} finally, the same approach also allows us to synthesize novel views of unseen human subjects (faces are blurred). The figure is best viewed in electronic format.}
    \label{fig:teaser}
\end{center}%


\begin{abstract}
Image-based volumetric humans using pixel-aligned features promise generalization to unseen poses and identities.
Prior work leverages global spatial encodings and multi-view geometric consistency to reduce spatial ambiguity.
However, global encodings often suffer from overfitting to the distribution of the training data, and it is difficult to learn multi-view consistent reconstruction from sparse views.
In this work, we investigate common issues with existing spatial encodings and propose a simple yet highly effective approach to modeling high-fidelity volumetric humans from sparse views.
One of the key ideas is to encode relative spatial 3D information via sparse 3D keypoints. 
This approach is robust to the sparsity of viewpoints and cross-dataset domain gap.
Our approach outperforms state-of-the-art methods for head reconstruction. 
On human body reconstruction for unseen subjects, we also achieve performance comparable to prior work that uses a parametric human body model and temporal feature aggregation. 
Our experiments show that a majority of errors in prior work stem from an inappropriate choice of spatial encoding and thus we suggest a new direction for high-fidelity image-based human modeling.
\end{abstract}

\section{Introduction}
\label{sec:intro}

\begin{figure}[t]
    \begin{center}
        \includegraphics[width=\textwidth]{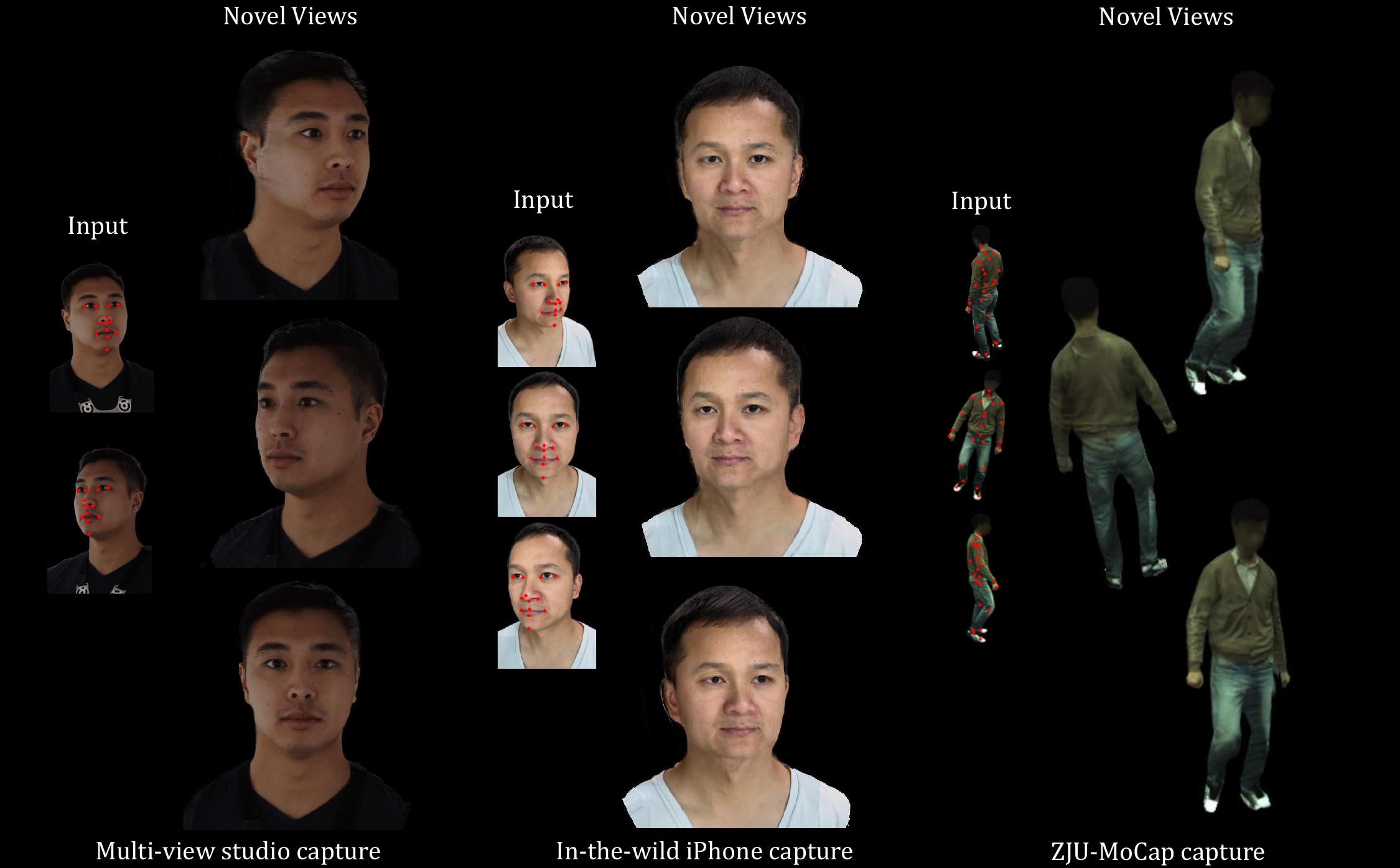}
    \end{center}
    \caption{\textbf{(a)} Our approach allows us to synthesize high-fidelity volumetric humans from two or three views of unseen subjects. \textbf{(b)} A model learned on studio captures can be used without modification on in-the-wild iPhone captures; and \textbf{(c)} finally, the same approach also allows us to synthesize novel views of unseen human subjects (faces are blurred). The figure is best viewed in electronic format.}
    \label{fig:teaser}
\end{figure}

3D renderable human representations are an important component for social telepresence, mixed reality, and a new generation of entertainment platforms.
Classical mesh-based methods require dense multi-view stereo~\cite{matusik2000image,vlasic2008articulated,vlasic2009dynamic} or depth sensor fusion~\cite{yu2018doublefusion}. The fidelity of these methods is limited due to the difficulty of accurate geometry reconstruction.
Recently, neural volumetric representations~\cite{lombardi2019nv,mildenhall2020nerf} have enabled high-fidelity human reconstruction, especially where accurate geometry is difficult to obtain (e.g. hair).
By injecting human-specific parametric shape models~\cite{blanz1999morphable,loper2015smpl}, extensive multi-view data capture can be reduced to sparse camera setups~\cite{peng2021neuralbody,gafni2021nerface}.
However, these learning-based approaches are subject-specific and require days of training for each individual subject, which severely limits their scalability. 
Democratizing digital volumetric humans requires an ability to instantly create a personalized reconstruction of a user from two or three snaps (from different views) taken from their phone. Towards this goal, we learn to generalize metrically accurate image-based volumetric humans from two or three views.

Fully convolutional pixel-aligned features utilizing multi-scale information have enabled better performance for various 2D computer vision tasks~\cite{bansal2017pixelnet,openpose,long2015fully,hourglass}, including the generalizable reconstruction of unseen subjects \cite{saito2019pifu,kwon2021neuralhumanperformer,raj2021pva,yu2021pixelnerf}
%
Pixel-aligned neural fields infer field quantities given a pixel location and spatial encoding function (to avoid ray-depth ambiguity). Different spatial encoding functions~\cite{gao2020portraitnerf,he2021arch++,huang2020arch,raj2021pva} have been proposed in the literature. However, the effect of spatial encoding is not fully understood. 
%
%
%
In this paper, we provide an extensive analysis of spatial encodings for modeling pixel-aligned neural radiance fields for human faces. 
Our experiments show that the choice of spatial encoding influences the reconstruction quality and generalization to novel identities and views.
The models that use camera depth overfit to the training distribution, and multi-view stereo constraints are less robust to sparse views with large baselines.

We present a simple yet highly effective approach based on sparse 3D keypoints to address the limitations of existing approaches. 
%
3D keypoints are easy to obtain using an off-the-shelf 2D keypoint detector~\cite{openpose} and a simple triangulation of multi-views~\cite{hartley2003multiple}.
We treat 3D keypoints as spatial anchors and encode relative 3D spatial information to those keypoints.
Unlike global spatial encoding~\cite{saito2019pifu,saito2020pifuhd,yu2021pixelnerf}, the relative spatial information is agnostic to camera parameters. This property allows the proposed approach to be robust to changes in pose.
3D keypoints also allow us to use the same approach for both human faces and bodies. 
%
Our approach achieves state-of-the-art performance for generating volumetric humans for unseen subjects from sparse-and-wide two or three views, and we can also incorporate more views to further improve performance. 
We also achieve performance comparable to Neural Human Performer (NHP)~\cite{kwon2021neuralhumanperformer} when it comes to full-body human reconstruction. NHP relies on accurate parametric body fitting and temporal feature aggregation, whereas our approach employs 3D keypoints alone. 
Our method is not biased~\cite{torralba2011unbiased} to the distribution of the training data. We can use the learned model (without modification) to never-before-seen iPhone captures.
We attribute our ability to generalize image-based volumetric humans to an unseen data distribution to our choice of spatial encoding.
Our key contributions include:
\begin{itemize}
    \item A simple approach that leverages sparse 3D keypoints and allows us to create high-fidelity state-of-the-art volumetric humans for unseen subjects from widely spread out two or three views.
    \item Extensive analysis on the use of spatial encodings to understand their limitations and the efficacy of the proposed approach.
    \item We demonstrate generalization to never-before-seen iPhone captures by training with only a studio-captured dataset. To our knowledge, no prior work has shown these results.
    
\end{itemize}

\section{Related Work}

Our goal is to create high-fidelity volumetric humans for unseen subjects from as few as two views.

\noindent\textbf{Classical Template-based Techniques:}
Early work on human reconstruction~\cite{ichim2015dynamic} required dense 3D reconstruction from a large number of images of the subject and non-rigid registration to align a template mesh to 3D point clouds.
Cao et al.~\cite{cao2016real} employ coarse geometry along with face blendshapes and a morphable hair model to address restrictions posed by dense 3D reconstruction.
Hu et al.~\cite{hu2017avatar} retrieve hair templates from a database and carefully compose facial and hair details.
Video Avatar~\cite{alldieck2018video} obtains a full-body avatar based on a monocular video captured using silhouette-based modeling.
The dependence on geometry and meshes restricts the applicability of these methods to faithfully reconstruct regions such as the hair, mouth, teeth, etc., where it is non-trivial to obtain accurate geometry.
%


\noindent\textbf{Neural Rendering:}
Neural rendering~\cite{tewari2020state,tewari2021advances} has tackled some of the challenges classical template-based approaches struggle with by directly learning components of the image formation process from raw sensor measurements.
2D neural rendering approaches \cite{kim2018deep, raj2021anr, prokudin2021smplpix, martin2018lookingood, meka2020deep, mihajlovic2021deepsurfels} employ surface rendering and a convolutional network to bridge the gap between rendered and real images.
The downside of these 2D techniques is that they struggle to synthesize novel viewpoints in a temporally coherent manner.
Deep Appearance Models~\cite{lombardi2018deep} employ a coarse 3D proxy mesh in combination with view-dependent texture mapping to learn personalized face avatars from dense multi-view supervision.
Using a 3D proxy mesh significantly helps with viewpoint generalization, but the approach faces challenges in synthesizing certain regions for which it is hard to obtain good 3D reconstruction, such as the hair and inside the mouth.
Current state-of-the-art methods such as NeuralVolumes~\cite{lombardi2019nv} and NeRF~\cite{mildenhall2020nerf} employ differentiable volumetric rendering instead of relying on meshes.
Due to their volumetric nature, these methods enable high-quality results even for regions where estimating 3D geometry is challenging.
Various extensions~\cite{lombardi2021mixture,wang2021hybridnerf,athar2021flameinnerf} have further improved quality. These methods require dense multi-view supervision for person-specific training and take 3--4 days to train a single model.

\noindent\textbf{Sparse View Reconstruction:}
Large scale deployment requires approaches that allow a user to take two or three pictures of themselves from multi-views and generate a digital human from this data.
The use of pixel-aligned features~\cite{saito2019pifu,saito2020pifuhd} further allows the use of large datasets for learning generalized models from sparse views.
Different approaches~\cite{chen2021mvsnerf,SRF,wang2021ibrnet,yu2021pixelnerf} have combined multi-view constraints and pixel-aligned features alongside NeRF to learn generalizable view-synthesis.
In this work, we observe that these approaches struggle to generate fine details given sparse views for unseen human faces and bodies.

\noindent\textbf{Learning Face and Body Reconstruction:}
Generalizable parametric mesh \cite{loper2015smpl,xu2020ghum,anguelov2005scape} and implicit \cite{mihajlovic2021leap,alldieck2021imghum,mihajlovic2022coap,wang2021metaavatar,xu2021hnerf} body models can provide additional constraints for learning details from sparse views. 
Recent approaches have incorporated priors specific to human faces~\cite{gafni2021nerface,zheng2022IMavatar,rebain2022lolnerf,grassal2022neural,buehler2021varitex,Chen2022authentic,wang2022faceverse} and human bodies~\cite{peng2021neuralbody,weng2022humannerf,zheng2021pamir,xu2021hnerf,peng2021animatable,zhao2022high,zheng2022structured,Wang:ARAH:ECCV2022} to reduce the dependence on multi-view captures.
Approaches such as H3DNet~\cite{ramon2021h3dnet} and SIDER~\cite{chatziagapi2021sider} use signed-distance functions (SDFs) for learning geometry priors from large datasets and perform test-time fine-tuning on the test subject.
PaMIR \cite{zheng2021pamir} uses volumetric features guided by a human body model for better generalization.
Neural Human Performer~\cite{kwon2021neuralhumanperformer} employs SMPL with pixel-aligned features and temporal feature aggregation.
In this work, we observe that the use of human 3D-keypoints provides necessary and sufficient constraints for learning from sparse-view inputs.
Our approach has high flexibility since it only relies on 3D keypoints alone and thus enables us to work both on human faces and bodies.
Prior methods have also employed various forms of spatial encoding for better learning.
For example, PVA~\cite{raj2021pva} and PortraitNeRF~\cite{gao2020portraitnerf} use face-centric coordinates. ARCH/ARCH++~\cite{huang2020arch,he2021arch++} use canonical body coordinates.
In this work, we extensively study the role of spatial encoding, and found that the use of a relative depth encoding using 3D keypoints leads to the best results.
Our findings enable us to learn a representation that generalizes to never-before-seen iPhone camera captures for unseen human faces.
In addition to achieving state-of-the-art results on volumetric face reconstruction from as few as two images, our approach can also be used for synthesizing novel views of unseen human bodies and achieves competitive performance to prior work~\cite{kwon2021neuralhumanperformer} in this setting.

\section{Preliminaries: Neural Radiance Fields} 
\myparagraph{NeRF}~\cite{mildenhall2020nerf} is a continuous function that represents the scene as a volumetric radiance field of color and density. 
Given a 3D point and a viewing direction $d \in \mathbb{R}^3$, NeRF estimates RGB color values and density $(c,\sigma)$ that are then accumulated via quadrature to calculate the expected color of each camera ray $r(t) = o + td$:
\begin{equation} \label{eq:vol_rendering}
    C(r) = \int_{t_n}^{t_f} \text{exp}\left(-\int_{t_n}^{t}\sigma(s)\,ds\right) \sigma(t) c(t, d)\,dt \,,
\end{equation}
where $t_n$ and $t_f$ define near and far bounds. 

\myparagraph{Pixel-aligned NeRF.} 
One of the core limitations of NeRF is that the approach requires per-scene optimization and does not work well for extremely sparse input views (e.g., two images).
To address these challenges, several recent methods \cite{yu2021pixelnerf,wang2021ibrnet,raj2021pva} propose to condition NeRF on pixel-aligned image features and generalize to novel scenes without retraining. 

\myparagraph{Spatial Encoding.} To avoid the ray-depth ambiguity, pixel-aligned neural fields \cite{saito2019pifu,saito2020pifuhd,raj2021pva,yu2021pixelnerf} attach spatial encoding to the pixel-aligned feature.  
PIFu~\cite{saito2019pifu} and related methods~\cite{saito2020pifuhd,yu2021pixelnerf} use depth value in the camera coordinate space as spatial encoding, while PVA~\cite{raj2021pva} uses coordinates relative to the head position.
However, we argue that such spatial encodings are global and sub-optimal for learning generalizable volumetric humans. 
In contrast, our proposed relative spatial encoding provides a localized context that enables better learning and is more robust to changes in human pose. 



\begin{figure*}[t!]
    \begin{center}
        \includegraphics[width=1.0\textwidth]{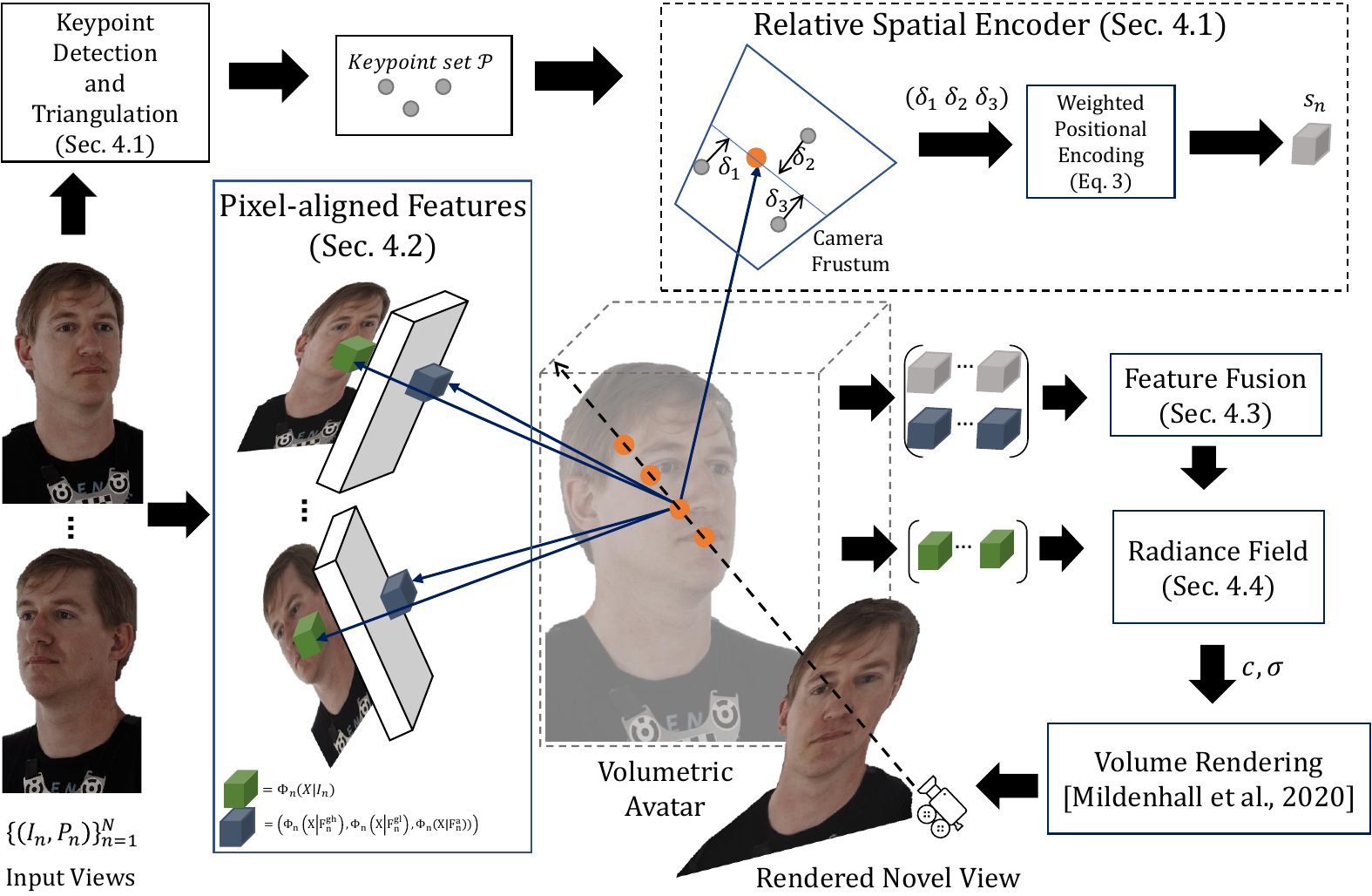}
    \end{center}
    \caption{
    \textbf{Overview.} 
    We learn a generalizable image-based neural radiance representation for volumetric humans. 
    Given a sparse set of input images $\{I_n\}_{n=1}^N$ and their respective camera parameters $P_n$, we first detect keypoints and lift them to 3D $\mathcal{P}$. The keypoints are used to provide the relative spatial encoding (Sec~\ref{subsec:sp_encoding}).
    The input images are simultaneously encoded via convolutional encoders and provide image-guided pixel-aligned features (Sec~\ref{subsec:pix_align}). 
    These two types of encoding are fused (Sec.~\ref{subsec:mv_fusion}) and condition the radiance field (Sec.~\ref{subsec:m_radiance_fields}). 
    The radiance field is decoupled by two MLPs, one that directly predicts view-independent density value $\sigma$, and the other one which predicts blending weights that are used to output the final color value $c$ by blending image pixel values $\{\Phi(X|I_n)\}_{n=1}^N$. 
    The predicted color and density values are accumulated along the ray via volume rendering \cite{mildenhall2019llff} to render the volumetric representation from novel views.
    The rendered example in the figure is an actual output of our method for the displayed two input images of an unseen subject. 
    }
    \label{fig:overview}
\end{figure*}

\section{KeypointNeRF}
Our method is based on a radiance field function:
\begin{equation}
    f\left(X, d| \{(I_n, P_n)\}_{n=1}^N\right) = (c, \sigma)
\end{equation}
that infers a color $c \in \mathbb{R}^3$ and a density $\sigma \in \mathbb{R}$ value for any point in 3D space given its position $X \in \mathbb{R}^3$ and its viewing direction $d \in \mathbb{R}^3$ as input. In addition, the function has access to the $N$ input images $I_n$ and their camera calibrations $P_n$.
We model the function $f$ as a neural network that consists of four main parts; 
a spatial keypoint encoder (Sec.~\ref{subsec:sp_encoding}), two convolutional image encoders that extract pixel-aligned features (Sec.~\ref{subsec:pix_align}), an MLP fusion network that aggregates multiple pixel-aligned features (Sec.~\ref{subsec:mv_fusion}), and two MLPs that predict density $\sigma$ and color values $c$ (Sec.~\ref{subsec:m_radiance_fields}). 
The high-level overview of our method is illustrated in \figurename~\ref{fig:overview} and in the following we further describe its components. 


\subsection{Relative Spatial Keypoint Encoding}\label{subsec:sp_encoding}
Our method first leverages an off-the-shelf keypoint regressor \cite{openpose} to extract $K$ 2D keypoints from at least two input views. 
Then these points are triangulated and lifted to 3D $\mathcal{P} = \{p_k \in \mathbb{R}^3\}_{k=1}^K$ by using the direct linear transformation algorithm \cite{hartley2003multiple}. 
To spatially encode the query point $X$, we first compute the depth values of the query point and all keypoints w.r.t each input view $z(p_k|P_n)$ by the 2D projection and then estimate their relative depth difference $\delta_n(p_k, X) = z(p_k|P_n) - z(X|P_n)$. 
We further employ positional encoding $\gamma(\cdot)$ \cite{mildenhall2020nerf} and the Gaussian exponential kernel to create the final relative spatial encoding for the query point $X$ as:
\begin{equation} \label{eq:sp_encoding}
    s_n(X|\mathcal{P}) =  \Big[\text{exp}\big(\frac{-l_2(p_k, X)^2}{2\alpha^2}\big) \gamma\big(\delta_n(p_k, X)\big)\Big]_{k=1}^K\,,
\end{equation}
where $\alpha$ is a fixed hyper-parameter that controls the impact of each keypoint. 
We set this value to 5cm for facial keypoints and to 10cm for the human skeleton. 


\subsection{Convolutional Pixel-aligned Features}\label{subsec:pix_align}
In addition to the spatial encoding $s_n(X)$, we extract pixel-aligned features for the query point $X$ by encoding the input images $I_n \in \mathbb{R}^{H \times W \times 3}$ using two convolutional encoders. 

\myparagraph{Image Encoders.}
The first image encoder uses a single HourGlass~\cite{hourglass} network that generates both deep low-resolution $F^{gl}_n \in \mathbb{R}^{H/8 \times W/8 \times 64}$ and shallow high-resolution $F^{gh}_n \in \mathbb{R}^{H/2 \times W/2 \times 8}$ feature maps. 
This network learns a geometric prior of humans and its output is used to condition the density estimation network. 
The second encoder is a convolutional network with residual connections~\cite{johnson2016perceptual} that encodes input images $F^{a}_n \in \mathbb{R}^{H/4 \times W/4 \times 8}$ and provides an alternative pathway for the appearance information which is independent of the density prediction branch in the spirit of DoubleField~\cite{shao2021doublefield}.
Please see the supplemental material for further architectural details. 

\myparagraph{Pixel-aligned Features.}
To compute the pixel-aligned features, we project the query point on each feature plane $x = \pi(X|P_n) \in \mathbb{R}^2$ and bi-linearly interpolate the grid values. 
We define this operation of computing the pixel-aligned features (2D projection and interpolation) by the operator $\Phi_n(X|F)$, where $F$ can represent any grid of vectors for the $n$th camera: $F^{gl}_n\,,F^{gh}_n\,,F^{a}_n\,,I_n$. 

\subsection{Multi-view Feature Fusion}\label{subsec:mv_fusion}
To model a multi-view consistent radiance field, we need to fuse the per-view spatial encodings $s_n$ (Eq.~\ref{eq:sp_encoding}) and the pixel-aligned features $\Phi_n$. 

The spatial encoding $s_n$ is first translated into a feature vector via a single-layer perceptron. 
This feature is then jointly blended with the deep low-resolution pixel-aligned feature $\Phi_n(X|F^{gl}_n)$ by a two-layer perceptron. 
The output is then concatenated with the shallow high-resolution feature $\Phi_n(X|F^{gh}_n)$ and processed by an additional two-layer perceptron that outputs per-view 64-dimensional feature vector that jointly encodes the blended spatial encoding and the pixel-aligned information. 
These multi-view features are then fused into a single feature vector $G_X \in \mathbb{R}^{128}$ by the mean and variance pooling operators as in~\cite{wang2021ibrnet}. 

\subsection{Modeling Radiance Fields} \label{subsec:m_radiance_fields}
The radiance field is modeled via decoupled MLPs for density $\sigma$ and color $c$ prediction. 

\myparagraph{Density Fields.} 
The density network is implemented as a four-layer MLP that takes as input the geometry feature vector $G_X$ and predicts the density value $\sigma$. 

\myparagraph{View-dependent Color Fields.} 
We implement an additional MLP to output the consistent color value $c$ for a given query point $X$ and its viewing direction $d$ by blending image pixel values $\{\Phi(X|P_n)\}_{n=1}^N$ similarly to IBRNet~\cite{wang2021ibrnet}. 
The input to this MLP is 
{\it 1)} the extracted geometry feature vector $G_X$ that ensures geometrically consistent renderings, 
{\it 2)} the additional pixel-aligned features $\Phi_n(X|F^{a}_n)$, 
{\it 3)} the corresponding pixel values $\Phi_n(X|I_n)$, and 
{\it 4)} the view direction that is encoded as the difference between the view direction $d$ and the camera views along with their dot product. 

These inputs are concatenated and augmented with the mean and variance vectors computed over the multi-view pixel-aligned features, and jointly propagated through a nine-layer perceptron with residual connections which predicts blending weights for each input view $\{w_n\}_{n=1}^N$.
These blending weights form the final color prediction by fusing the corresponding pixel-aligned color values:
\begin{equation} \label{eq:color_pred}
    c = \sum_{n=1}^N \frac{\text{exp}(w_n)\Phi_n(X|I_n)}{\sum_{i=1}^N \text{exp}(w_i)} \,.
\end{equation}

\section{Novel View Synthesis}
Given our radiance field function $f(X, d) = (c,\sigma)$, we render novel views via the volume rendering equation (\ref{eq:vol_rendering}), in which we define the near and far bound by analytically computing the intersection of the pixel ray and a geometric proxy that over-approximates the volumetric human and use the entrance and exit points as near and far bounds respectively.
For the experiments on human heads, we use a sphere with a radius of 30 centimeters centered around the keypoints, while for the human bodies we follow the prior work \cite{kwon2021neuralhumanperformer,peng2021neuralbody} and use a 3D bounding box.
Similar to NeRF~\cite{mildenhall2020nerf}, we employ a coarse-to-fine rendering strategy, but we employ the same network weights for both levels.

\subsection{Training and Implementation Details}
To train our network, we render $H^\prime \times W^\prime$ patches (as in \cite{schwarz2020graf}) by accumulating color and density values for 64 sampled points along the ray for the coarse and 64 more for the fine rendering. 
Our method is trained end-to-end by minimizing the mean $\ell_1$-distance between the rendered and the ground truth pixel values and the VGG~\cite{simonyan2014very}-loss applied over rendered and ground truth image patches:
\begin{equation}
    \mathcal{L} = \mathcal{L}_{\text{RGB}} + \mathcal{L}_{\text{VGG}}\,.
\end{equation}
The use of the VGG loss for NeRF training was also leveraged by the concurrent methods \cite{weng2022humannerf,alldieck2022photorealistic} to better capture high-frequency details. 
The final loss $\mathcal{L}$ is minimized by the Adam optimizer~\cite{kingma2014adam} with a learning rate of $1e^{-4}$ and a batch size of one. For the other parameters, we use their defaults. 
The background from all training and test input images is removed via an off-the-shelf matting network~\cite{MODNet}. 
Additionally for more temporally coherent novel-view synthesis at inference time, we clip the maximum of the dot product (introduced in Sec.~\ref{subsec:m_radiance_fields}) to 0.8 when the number of input images is two in the supplementary video. 

\section{Experiments}
In this section, we validate our method on three different reconstruction tasks and datasets:
{\it 1)} reconstruction of human heads from images captured in a multi-camera studio,
{\it 2)} reconstruction of human heads from in-the-wild images taken with the iPhone's camera, and
{\it 3)} reconstruction of human bodies on the public ZJU-MoCap dataset \cite{peng2021neuralbody}.
As evaluation metrics, we follow prior work \cite{wang2021ibrnet,raj2021pva,kwon2021neuralhumanperformer} and report the standard SSIM and PSNR metrics.

\subsection{Reconstruction of Human Heads from Studio Data} \label{subsec:exp_facedome}
\myparagraph{Dataset and Experimental Setup.}
Our captured data consists of 29 $1280 \times 768$-resolution cameras positioned in front of subjects.
We use a total of 351 identities and 26 cameras for training and 38 novel identities for evaluation.
At inference time, we reconstruct humans only from 2--3 input views.

\myparagraph{Baselines.}
As baseline, we employ the current state-of-the-art model IBRNet~\cite{wang2021ibrnet}. In addition, we add several other baselines by varying different types of encoding for the query points in our proposed reconstruction pipeline. 
Specifically, 
{\it 1)} our pipeline without any encoding, 
{\it 2)} with the camera $z$ encoding used in \cite{saito2019pifu,saito2020pifuhd}, 
{\it 3)} with the encoding of $xyz$ coordinates in the canonical space of a human head that is used in \cite{raj2021pva}, 
{\it 4)} relative encoding of $xyz$ as the distance between the query point and estimated keypoints, 
{\it 5)} our relative spatial encoding without distance weighing ($\alpha \to \infty$ in Eq.~\ref{eq:sp_encoding}), and 
{\it 6)} the proposed weighted relative encoding as described in the method section (Sec.~\ref{subsec:mv_fusion}). 
The last three models use a total of 13 facial keypoints that are visualized in \figurename~\ref{fig:teaser}. 
All methods are trained with a batch size of one for 150k training steps, except IBRNet~\cite{wang2021ibrnet} which was trained for 200k iterations. 
For more comparisons and baselines we refer the reader to the supplementary material and video. 
\begin{figure*}[h!]
    \scriptsize
    \setlength{\tabcolsep}{0.6mm} 
    \newcommand{\sz}{0.23}  
    \begin{tabular}{ccccc}  
           \scriptsize{\rotatebox{90}{\phantom{++++}Inputs}} &
           \includegraphics[width=\sz\linewidth]{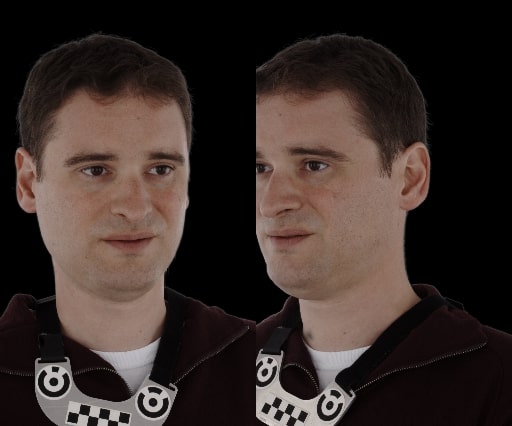} & 
           \includegraphics[width=\sz\linewidth]{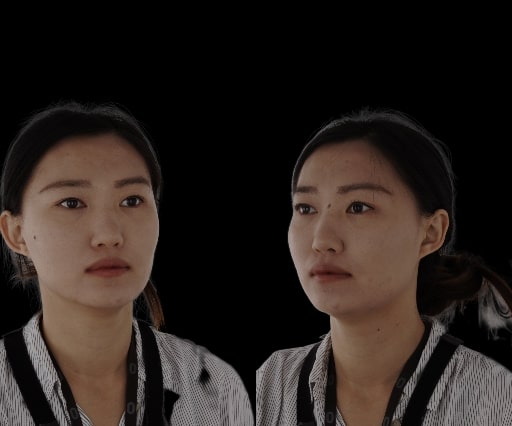} & 
           \includegraphics[width=\sz\linewidth]{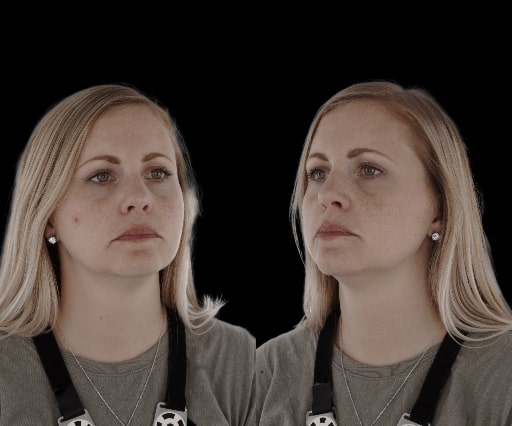} & 
           \includegraphics[width=\sz\linewidth]{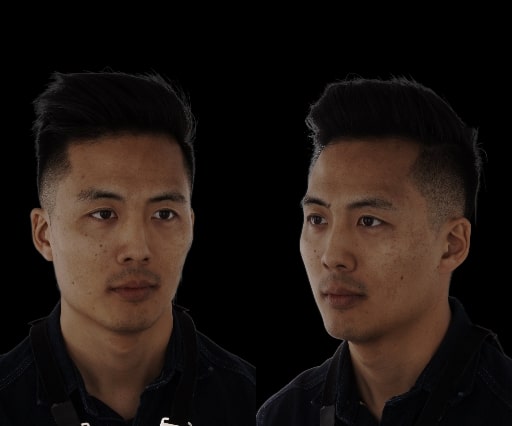} \\

            \scriptsize{\rotatebox{90}{\phantom{++++++}IBRNet~\cite{wang2021ibrnet}}}  &
            \includegraphics[width=\sz\linewidth, trim={0 200 0 50},clip]{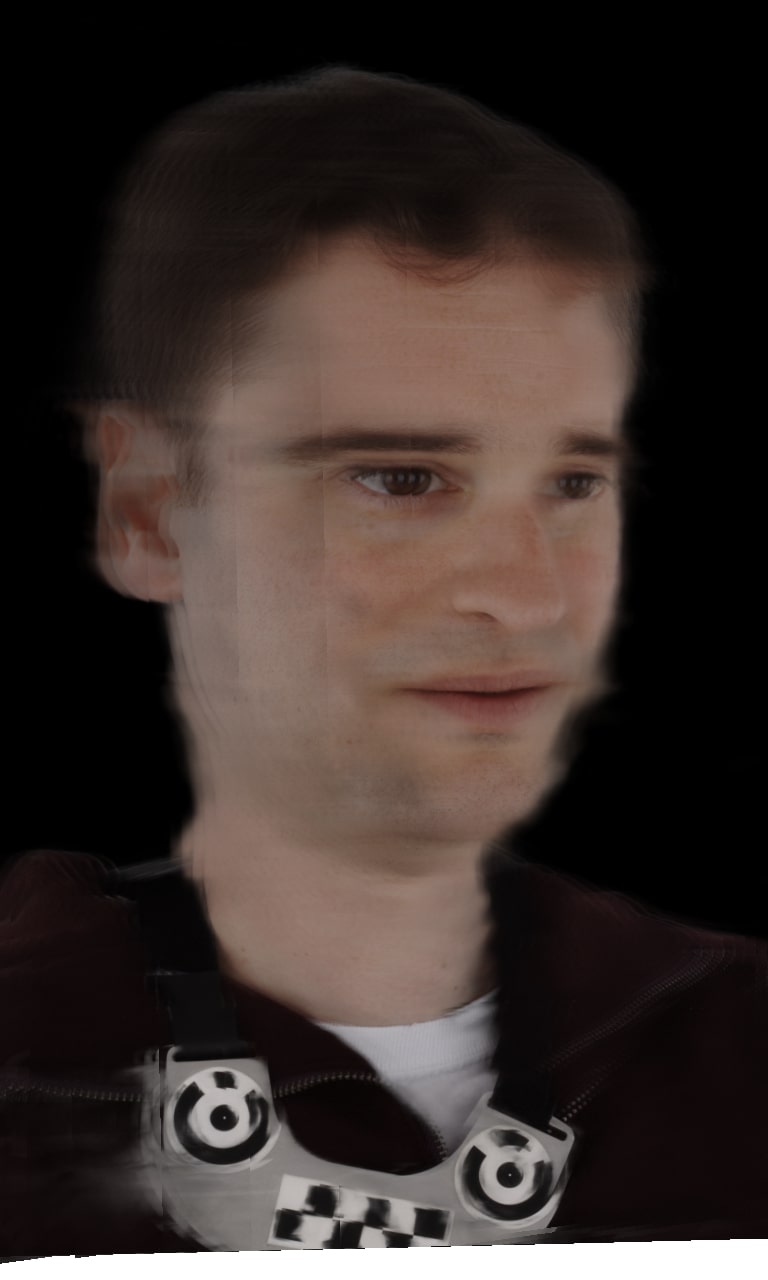} &
            \includegraphics[width=\sz\linewidth, trim={0 50 0 200},clip]{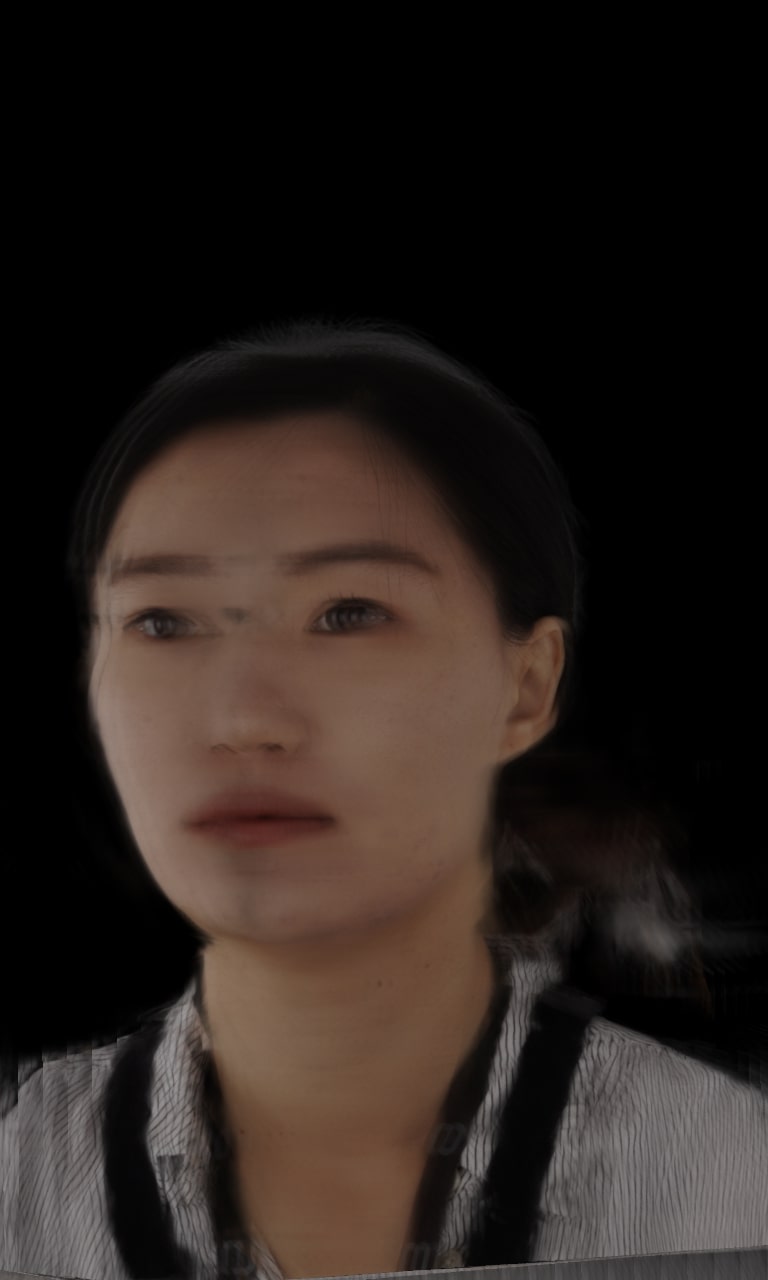} &
            \includegraphics[width=\sz\linewidth, trim={0 100 0 150},clip]{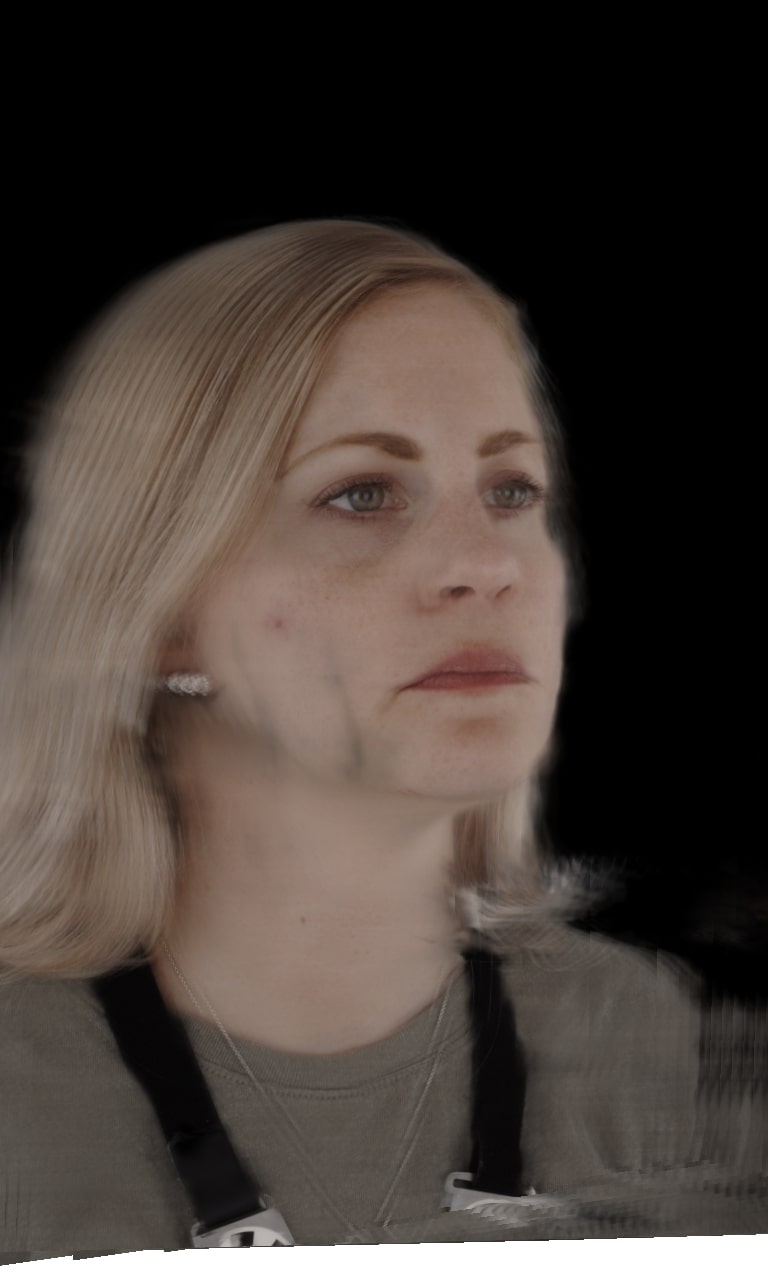} &
            \includegraphics[width=\sz\linewidth, trim={0 100 0 150},clip]{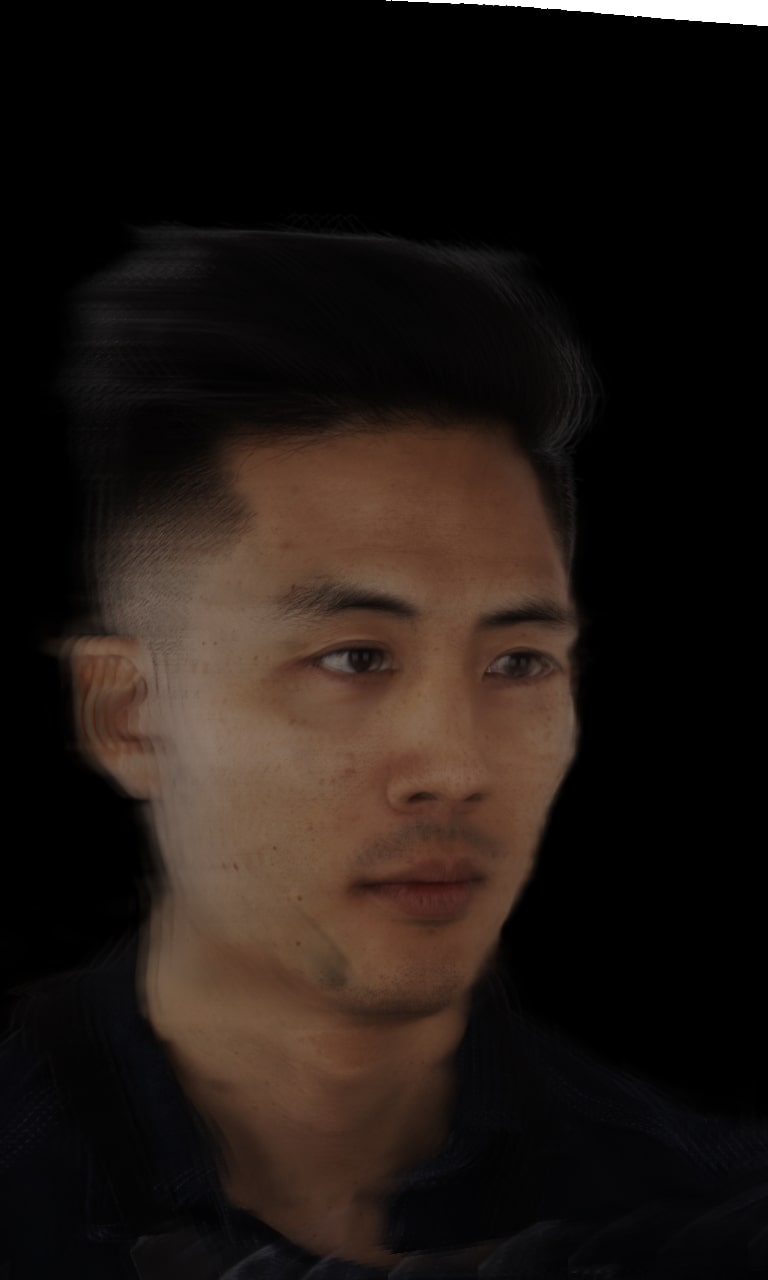} \\

            \scriptsize{\rotatebox{90}{\phantom{+++++}KeypointNeRF}}  &
            \includegraphics[width=\sz\linewidth, trim={0 200 0 50},clip]{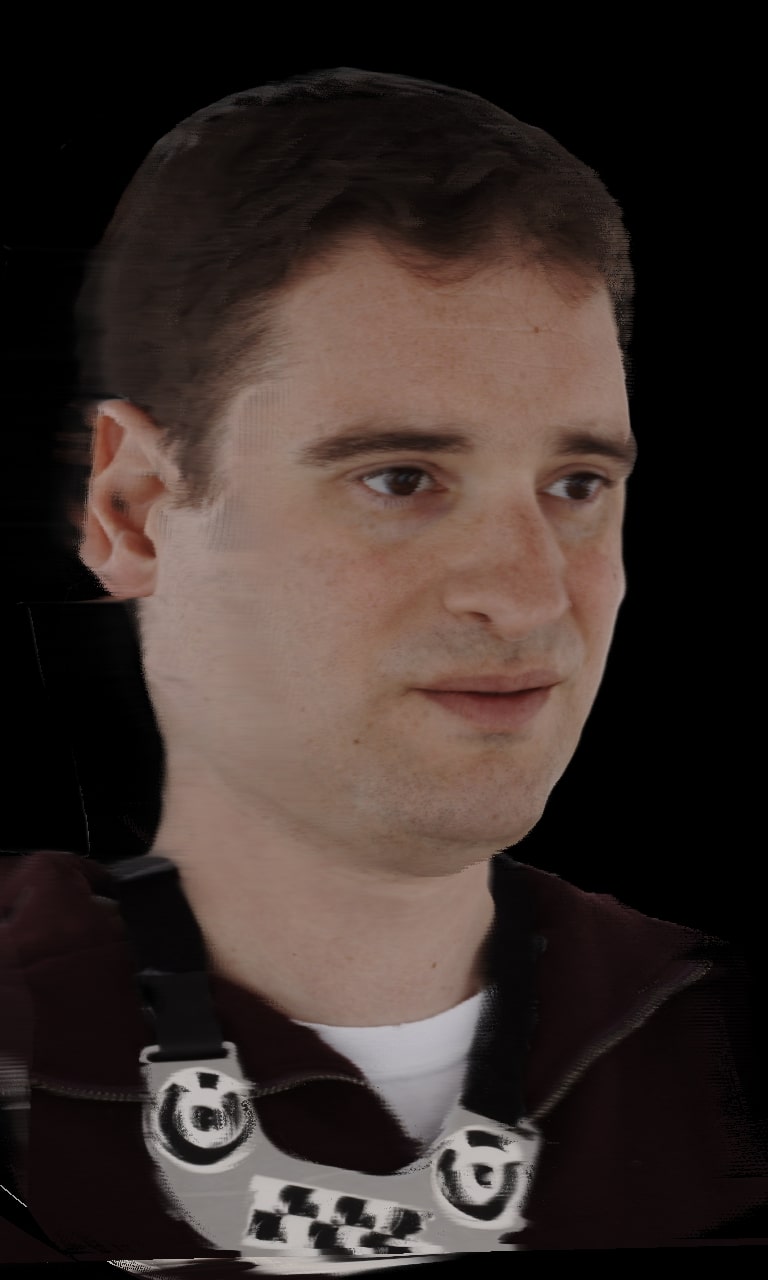} &
            \includegraphics[width=\sz\linewidth, trim={0 50 0 200},clip]{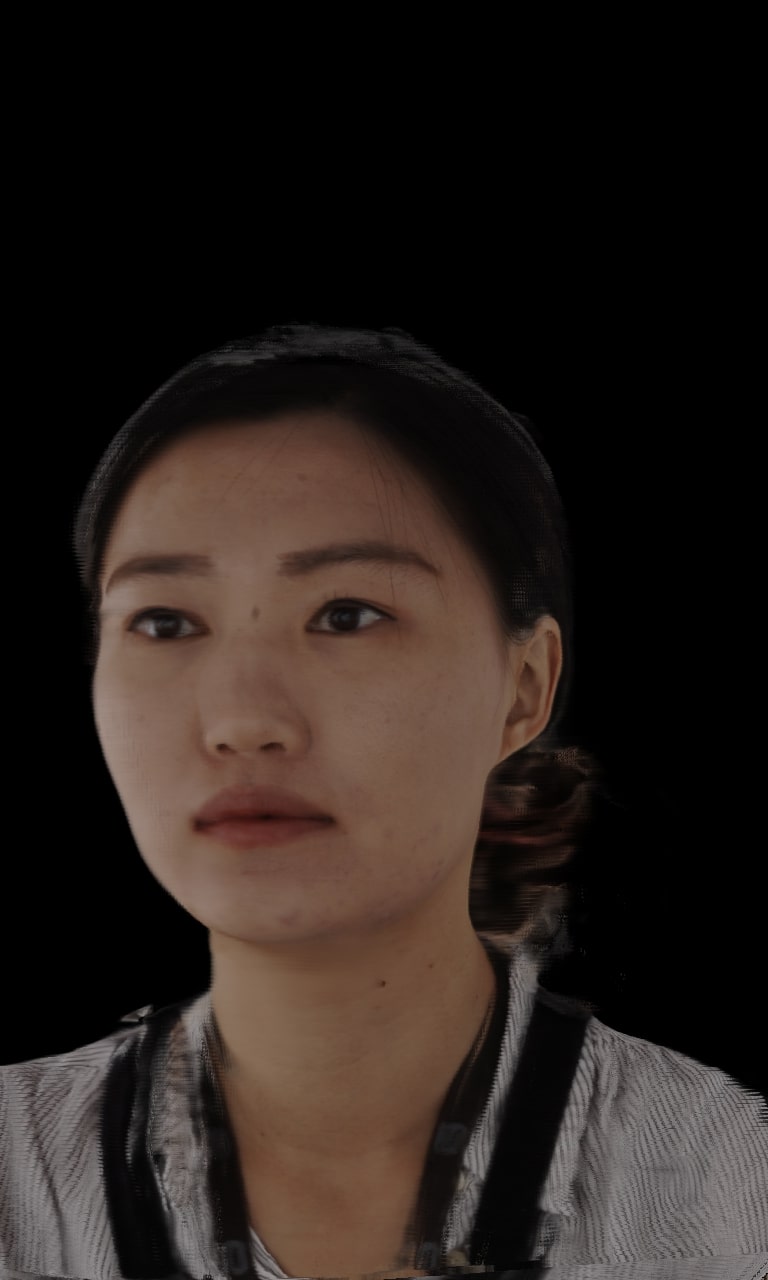} &
            \includegraphics[width=\sz\linewidth, trim={0 100 0 150},clip]{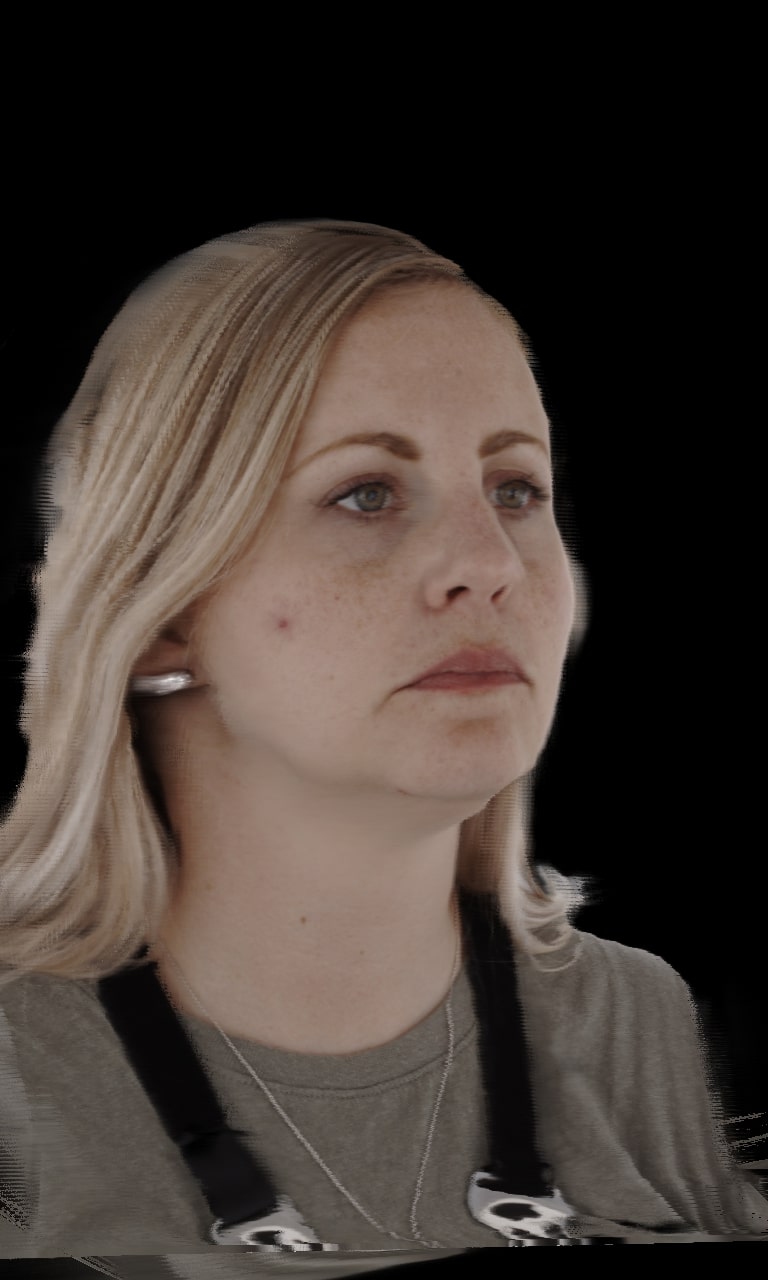} &
            \includegraphics[width=\sz\linewidth, trim={0 100 0 150},clip]{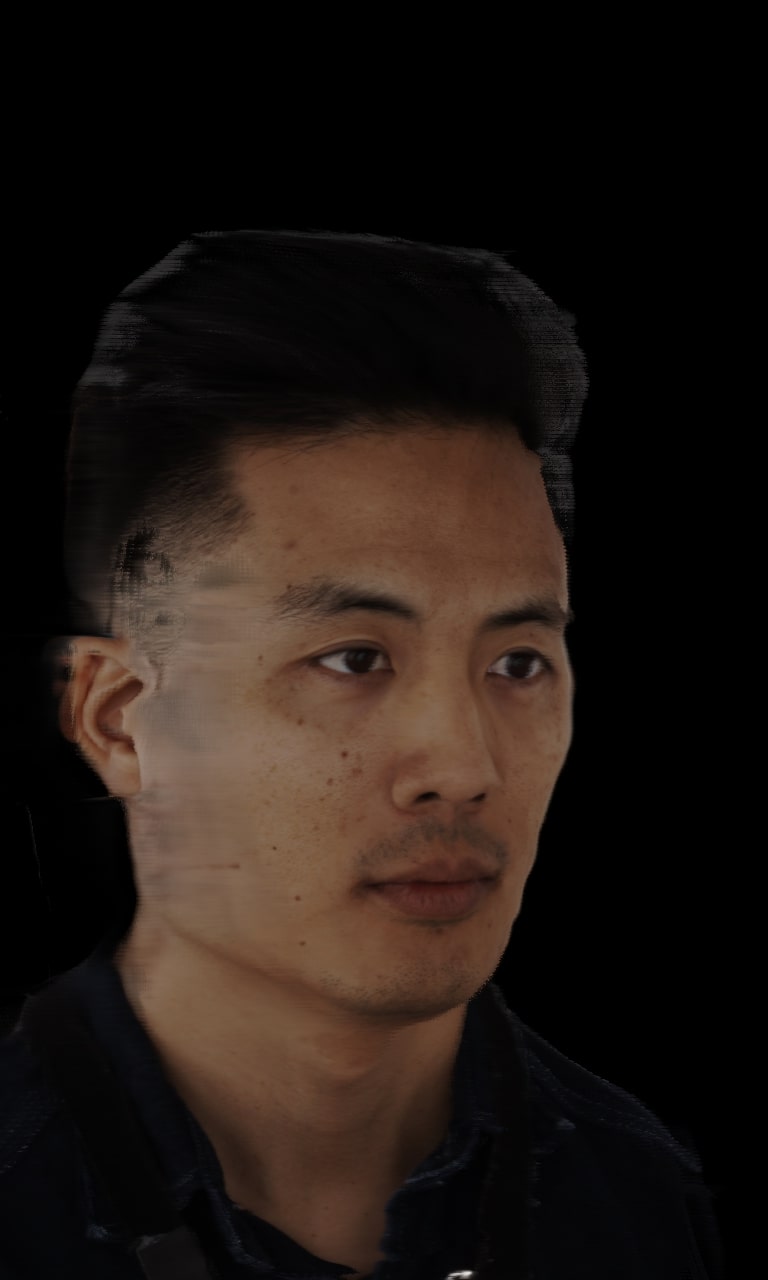} \\

            \scriptsize{\rotatebox{90}{\phantom{++++++++}GT}}  &
            \includegraphics[width=\sz\linewidth, trim={0 200 0 50},clip]{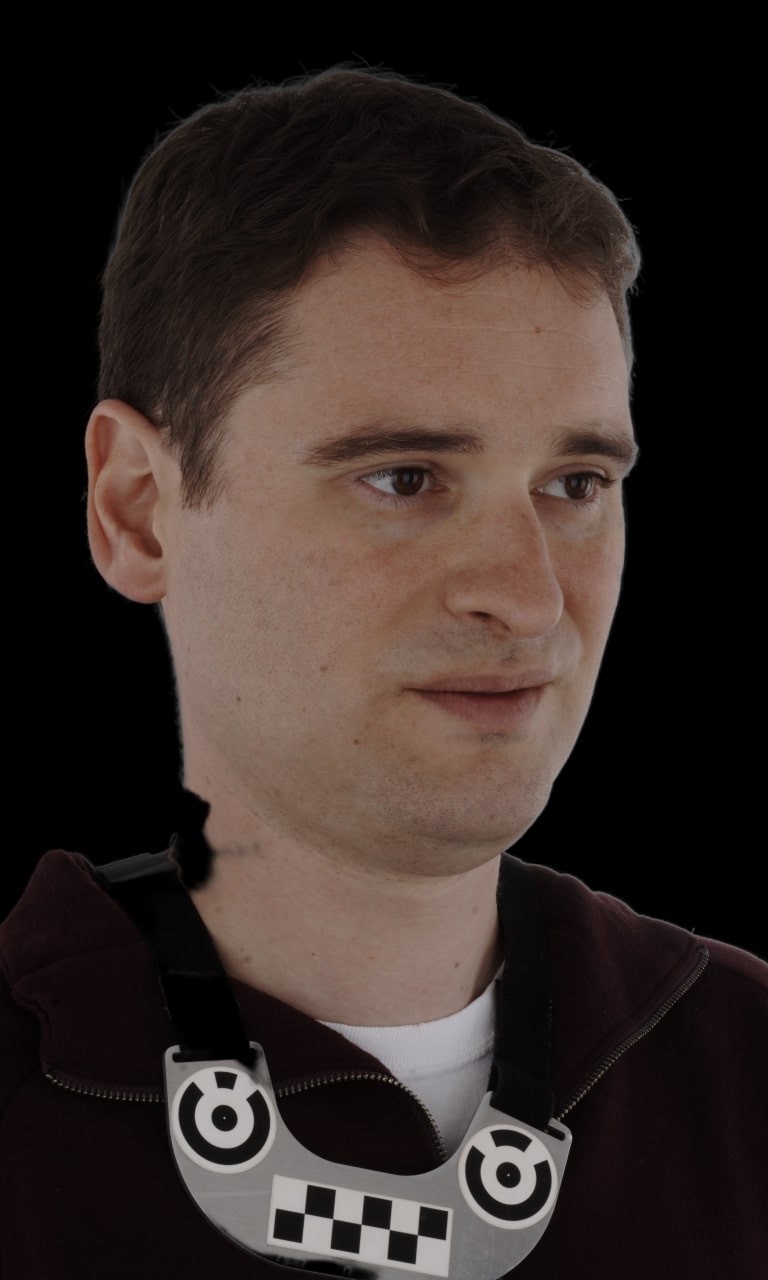} &
            \includegraphics[width=\sz\linewidth, trim={0 50 0 200},clip]{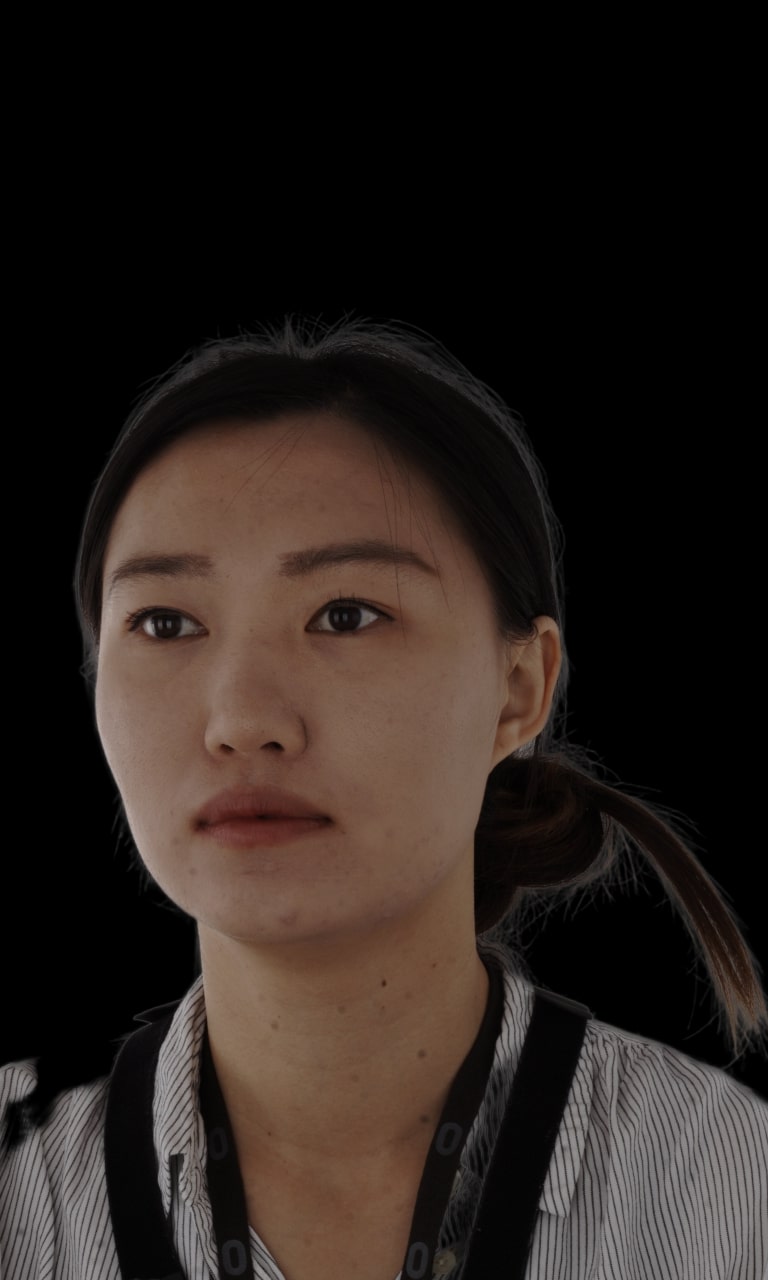} &
            \includegraphics[width=\sz\linewidth, trim={0 100 0 150},clip]{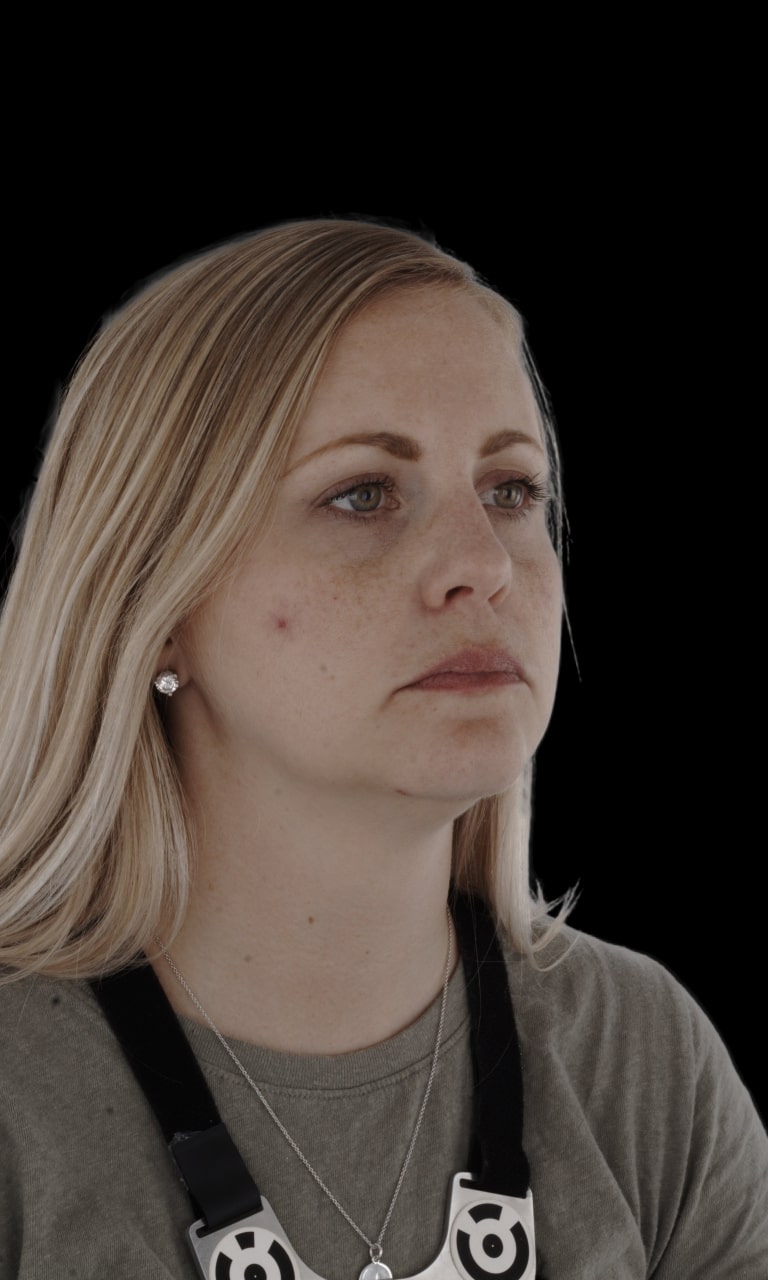} &
            \includegraphics[width=\sz\linewidth, trim={0 100 0 150},clip]{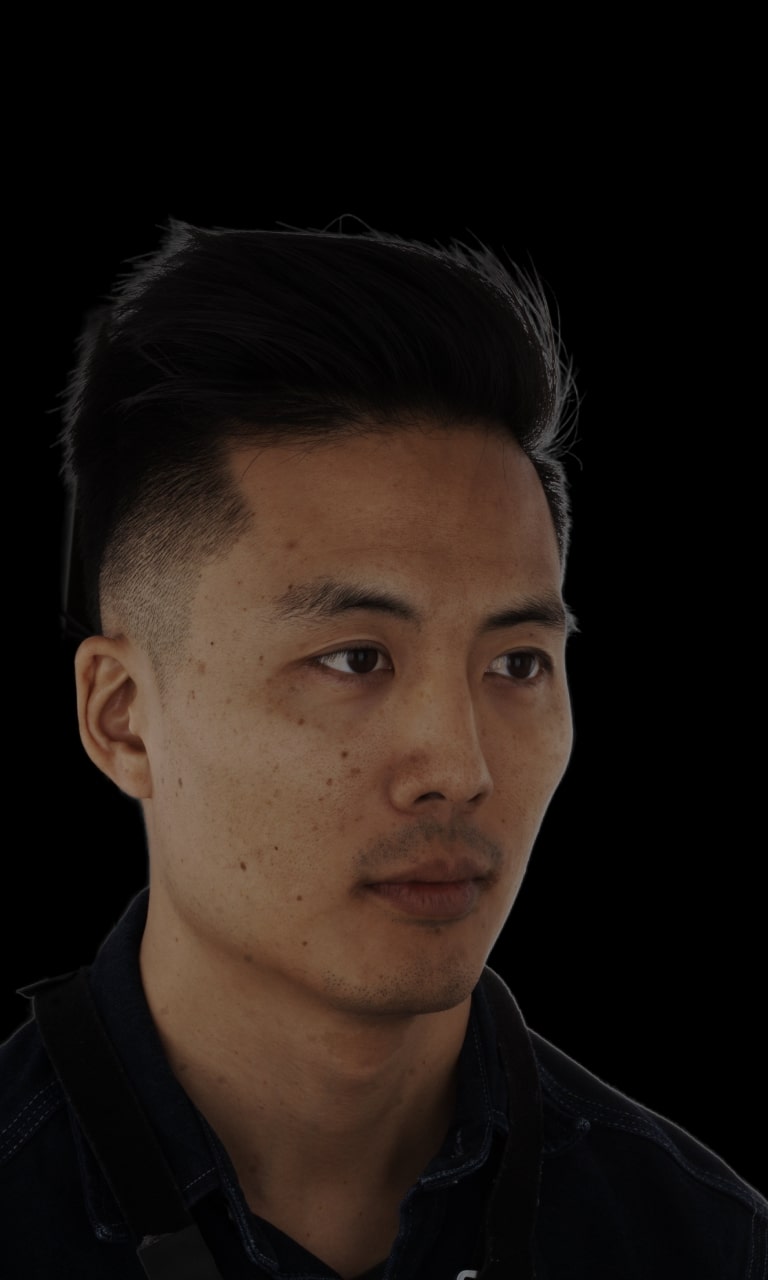} \\
    \end{tabular}
    \caption{\textbf{Studio Capture Results.} Reconstruction results on held-out subjects from only two input views. Best viewed in electronic format.}
    \label{fig:results_2views}
\end{figure*}
\begin{table*}
    \centering
    \scriptsize
    \setlength{\tabcolsep}{1.5pt}
    \caption{\textbf{Studio Capture Results.} Quantitative comparison with IBRNet~\cite{wang2021ibrnet} and different types of spatial encoding. Visual results are provided in \figurename~\ref{fig:results_2views}}
    \label{tab:abl_encodings}
    \begin{tabular}{@{}clcc@{}}
        \toprule
        &                                               & SSIM$\uparrow$ & PSNR$\uparrow$ \\
        \midrule
        & PVA~\cite{raj2021pva}                                                 &   81.95                               &   25.87 \\
        & IBRNet~\cite{wang2021ibrnet}                                          &   82.39                               &   27.14 \\\hline
        \multirow{6}{*}{\rotatebox{90}{Our Pipeline}}
        & 1) no encoding	                                                    &	84.38	                            &	27.16 \\
        & 2) camera z encoding~\cite{saito2019pifu}	                            &	77.86	                            &	22.66 \\
        & 3) canonical xyz encoding~\cite{raj2021pva}                           &	83.11	                            &	26.33 \\
        & 4) relative xyz encoding	                                            &	83.66	                            &	27.05 \\
        & 5) relative z encoding	                                            &	\cellcolor[RGB]{\colorsecond}84.72	&	\cellcolor[RGB]{\colorfirst}27.66  \\
        & 6) KeypointNeRF (distance weighted relative z encoding, Eq.~\ref{eq:sp_encoding})   &	\cellcolor[RGB]{\colorfirst}85.19   &	\cellcolor[RGB]{\colorsecond}27.64 \\
        \bottomrule\\
    \end{tabular}
\end{table*}
\myparagraph{Results.} 
We provide novel view synthesis (Fig.~\ref{fig:results_2views}) results for unseen identities that have been reconstructed from only two input images.
The results clearly demonstrate that the rendered images of our method are significantly sharper compared to the baselines and are of significantly higher quality.
This improvement is confirmed by the quantitative evaluation (Tab.~\ref{tab:abl_encodings}) which further indicates that the proposed distance weighting of the relative spatial encoding improves the reconstruction quality. 
The third-best performing method is our pipeline without any spatial encoding. 
However, such a method does not generalize well to other capture systems as we will demonstrate in the next section on in-the-wild captured data. 
%
\begin{table*}[t!]
    \centering
    \scriptsize
    
    \setlength{\tabcolsep}{6.0pt}
    \caption{\textbf{Robustness to Different Noise Levels.} Perturbing our keypoints vs.~perturbing head position in the canonical xyz encoding used in \cite{raj2021pva}. Our proposed encoding demonstrates significantly slower performance degradation as the noise increases }
    \label{tab:noise}
    \begin{tabular}{@{}ccc|cc@{}}
        \toprule
        Noise level        & \multicolumn{2}{c}{Canonical xyz encoding}              & \multicolumn{2}{c}{Our keypoint encoding}             \\

        [{\it mm}]          & SSIM$\uparrow$ & PSNR$\uparrow$ & SSIM$\uparrow$ & PSNR$\uparrow$ \\
        \midrule
no noise	            &	83.65	&	27.05	&	85.19	&	27.64\\
\phantom{0}1	        &	82.79	&	26.24	&	85.20	&	27.64\\
\phantom{0}2	        &	82.26	&	26.05	&	85.08	&	27.62\\
\phantom{0}3	        &	81.58	&	25.48	&	84.96	&	27.59\\
\phantom{0}4	        &	80.92	&	25.08	&	84.95	&	27.56\\
\phantom{0}5	        &	80.36	&	25.16	&	84.80	&	27.51\\
10 	                    &	76.62	&	22.23	&	83.69	&	27.10\\
15 	                    &	73.33	&	20.26	&	82.33	&	26.47\\
20 	                    &	70.40	&	18.87	&	81.17	&	25.77\\
        \bottomrule\\
    \end{tabular}
\end{table*}

\myparagraph{Robustness to Different Noise Levels.}
We evaluate the robustness of our relative spatial encoding and the encoding in canonical space proposed by Raj et al.~\cite{raj2021pva} by adding different noise levels to the estimated keypoints and the head center respectively.
The results reported in Tab.~\ref{tab:noise} show that our proposed encoding based on keypoints is significantly more robust compared to modeling in an object-specific canonical space.
Note that the canonical encoding requires head template fitting for which we used the ground truth estimation from all views and which, in practice, is very erroneous or even infeasible from two views alone. 

\myparagraph{Dynamic scenes.}
Although our model is trained only with a neutral face, it generalizes well to dynamic expressions and outperforms the baseline methods.
We evaluate the trained models on $38$ test subjects performing eight different expressions and report results in Tab.~\ref{tab_fig:dynamic_faces}. 

\begin{table}[!ht]
    \caption{\textbf{Dynamic expressions.} Our model is more accurate than the baselines}
    \begin{minipage}{.65\textwidth} %
    \begin{center}
        \includegraphics[width=\textwidth]{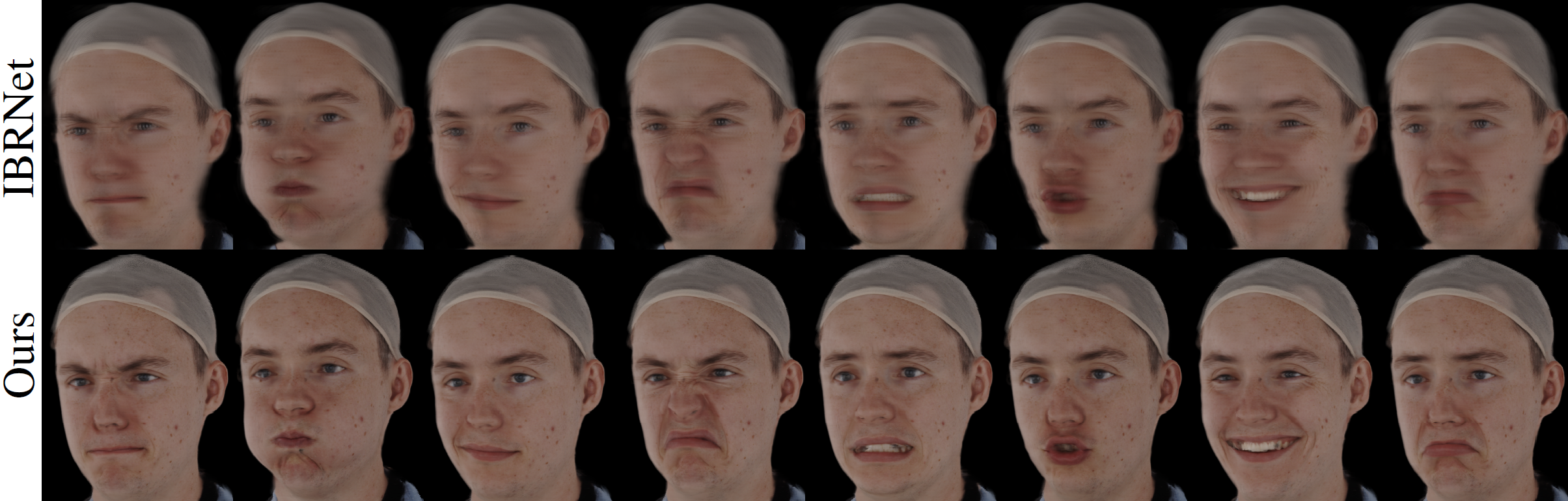}
    \end{center}

    \end{minipage} %
    \begin{minipage}{.2\textwidth} %
        \centering
        \scriptsize
        \setlength{\tabcolsep}{0.2pt}
        \begin{tabular}{@{}lccc@{}}
            \toprule
                                                                & SSIM$\uparrow$ & PSNR$\uparrow$\\
            \midrule
            IBRNet~\cite{wang2021ibrnet}                        &   82.64   &   26.79 \\
            Ours (no keypoints)	                             &   84.97   &   27.14 \\
            Ours	                                            &   \textbf{85.31}   &   \textbf{27.30} \\
            \bottomrule\\
        \end{tabular}
        \label{tab_fig:dynamic_faces}
    \end{minipage}
\end{table}

\subsection{Reconstruction from In-the-wild Captures}\label{subsec:iphone_results}
\begin{figure*}
    \scriptsize
    \setlength{\tabcolsep}{0.6mm} 
    \newcommand{\sz}{0.23}  
    \begin{tabular}{ccccc}  
           \scriptsize{\rotatebox{90}{\phantom{++}Inputs}} & 
           \includegraphics[width=\sz\linewidth, trim={0 10 0 30},clip]{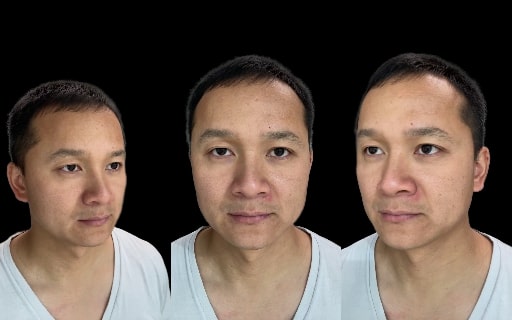} & 
           \includegraphics[width=\sz\linewidth, trim={0 10 0 30},clip]{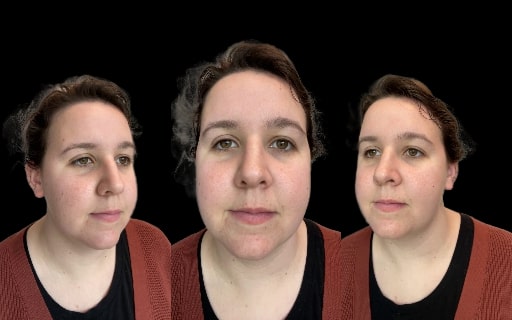} & 
           \includegraphics[width=\sz\linewidth, trim={0 10 0 30},clip]{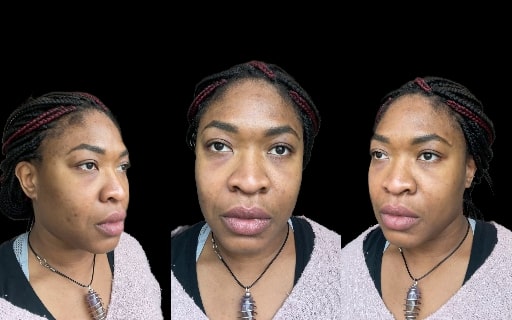} & 
           \includegraphics[width=\sz\linewidth, trim={0 10 0 30},clip]{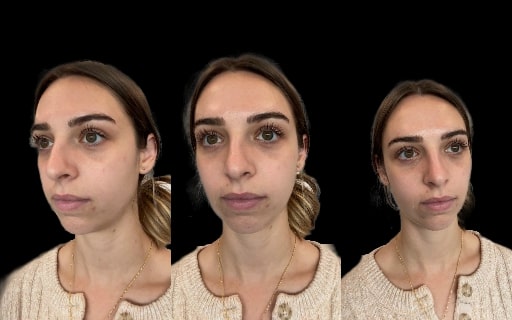} \\

            \scriptsize{\rotatebox{90}{\phantom{++++++}IBRNet~\cite{wang2021ibrnet}}}  &
            \includegraphics[width=\sz\linewidth, trim={0 150 0 300},clip]{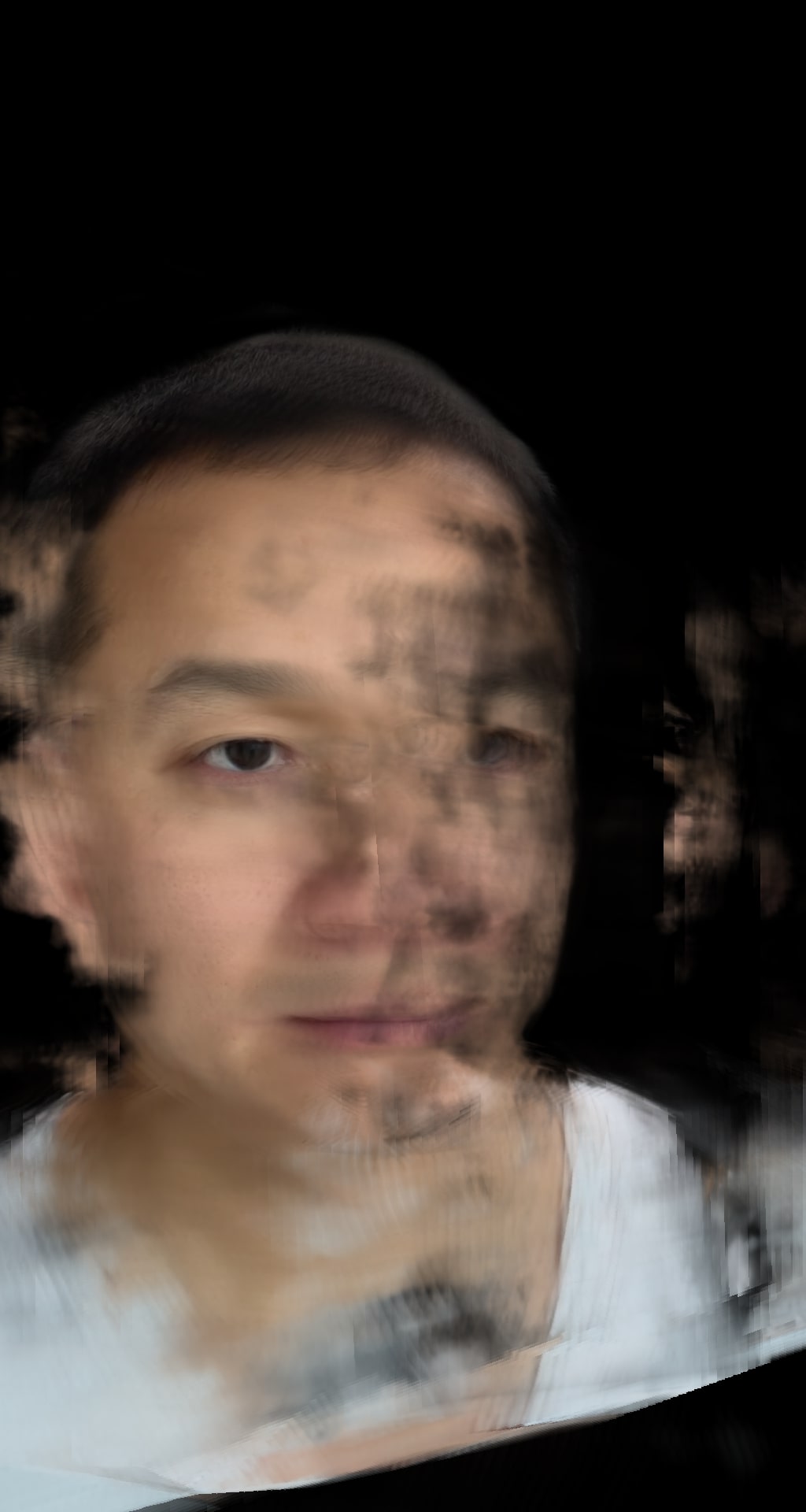} &
            \includegraphics[width=\sz\linewidth, trim={0 150 0 300},clip]{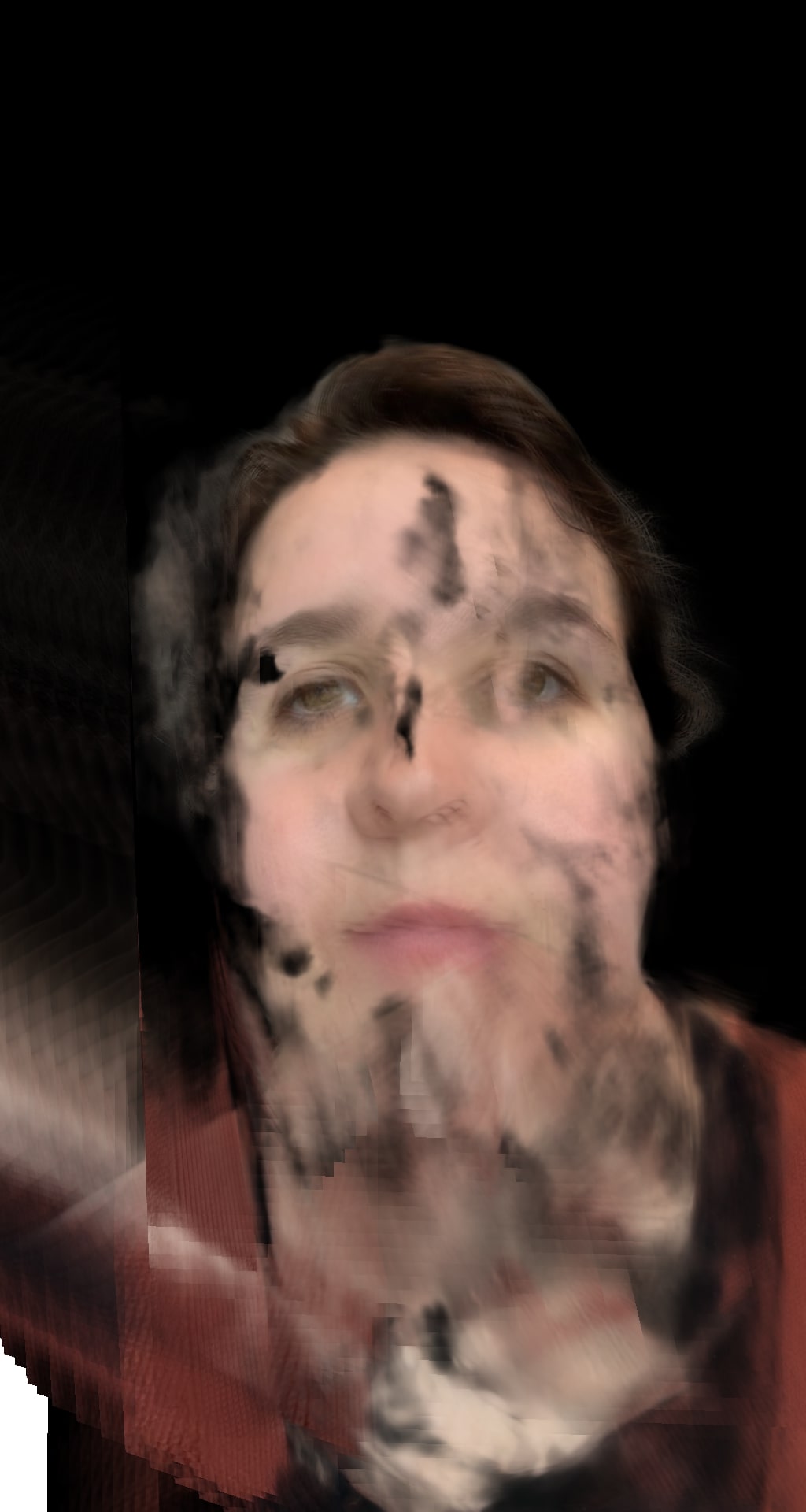} &
            \includegraphics[width=\sz\linewidth, trim={0 150 0 300},clip]{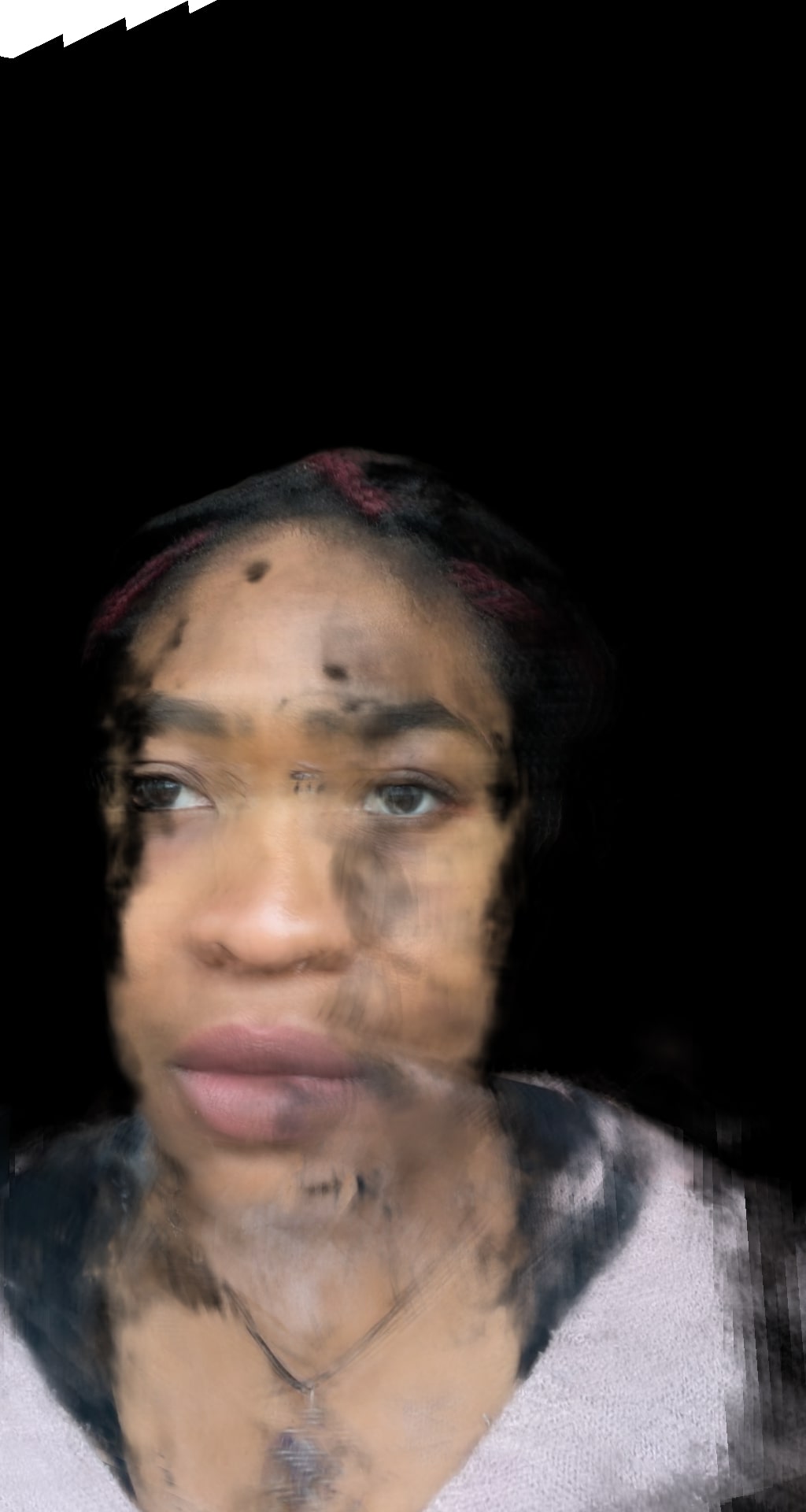} &
            \includegraphics[width=\sz\linewidth, trim={0 150 0 300},clip]{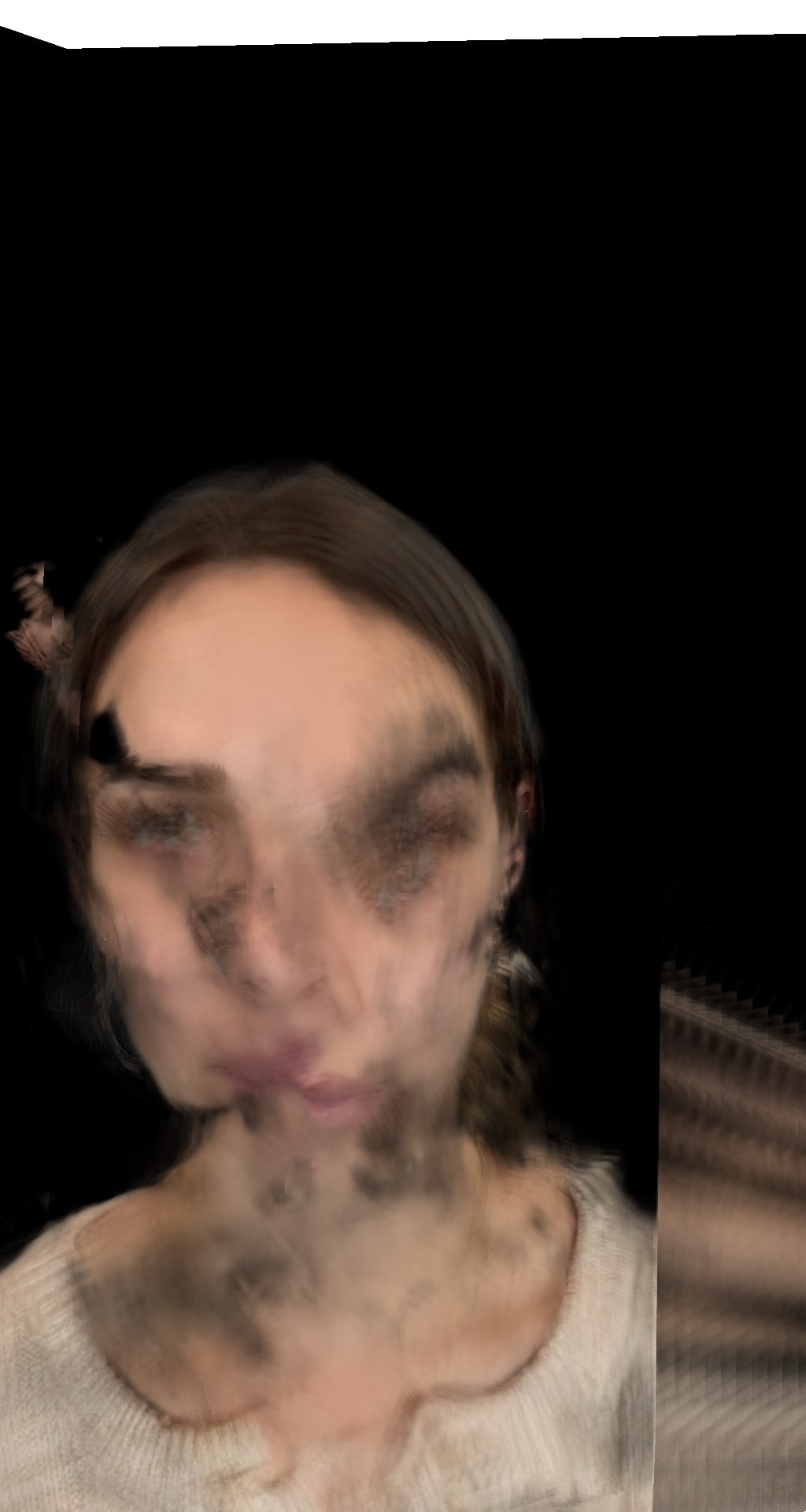} \\

            \scriptsize{\rotatebox{90}{\phantom{+++} Ours (no keypoints)}}  &
            \includegraphics[width=\sz\linewidth, trim={0 150 0 300},clip]{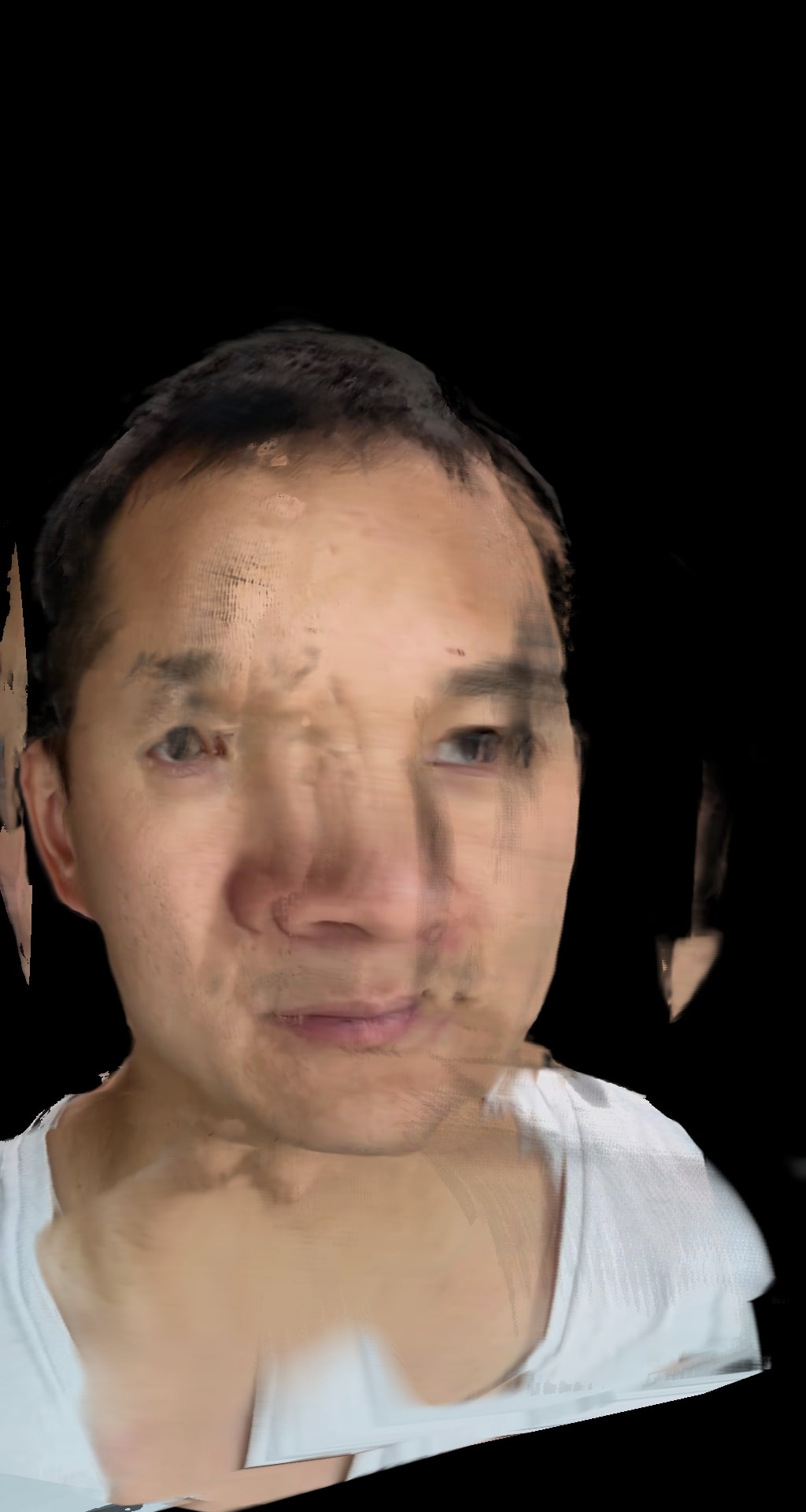} &
            \includegraphics[width=\sz\linewidth, trim={0 150 0 300},clip]{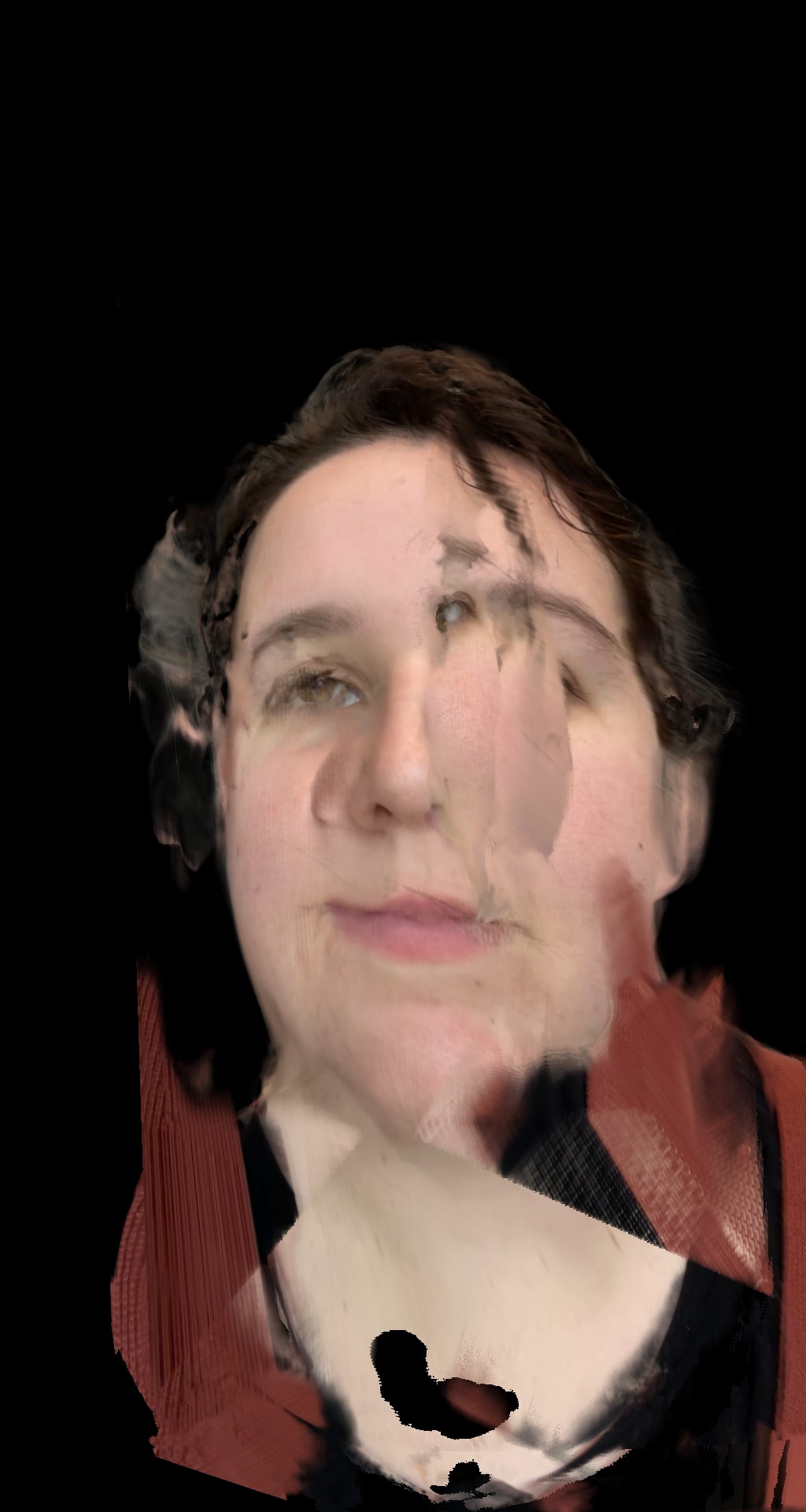} &
            \includegraphics[width=\sz\linewidth, trim={0 150 0 300},clip]{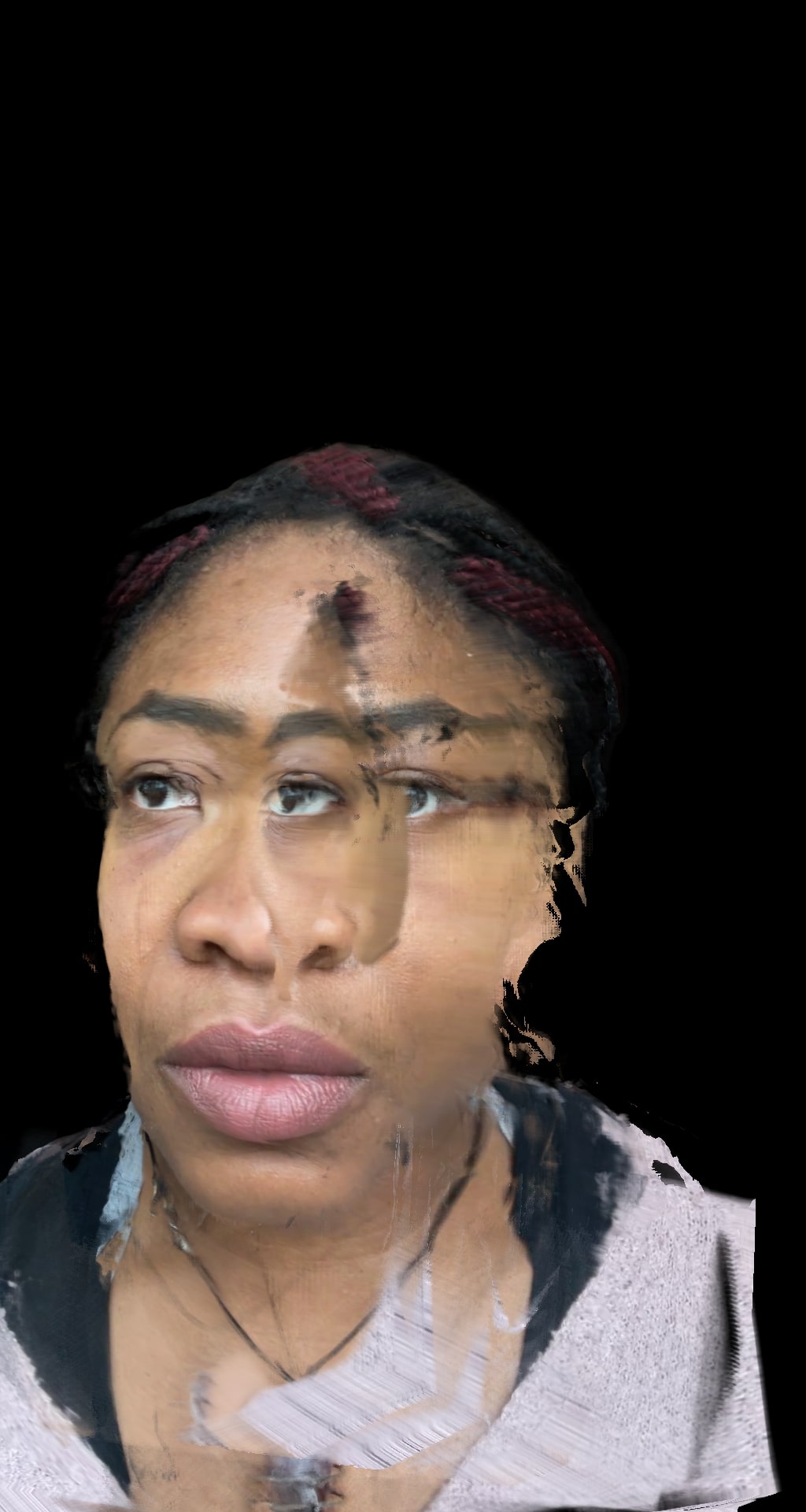} &
            \includegraphics[width=\sz\linewidth, trim={0 150 0 300},clip]{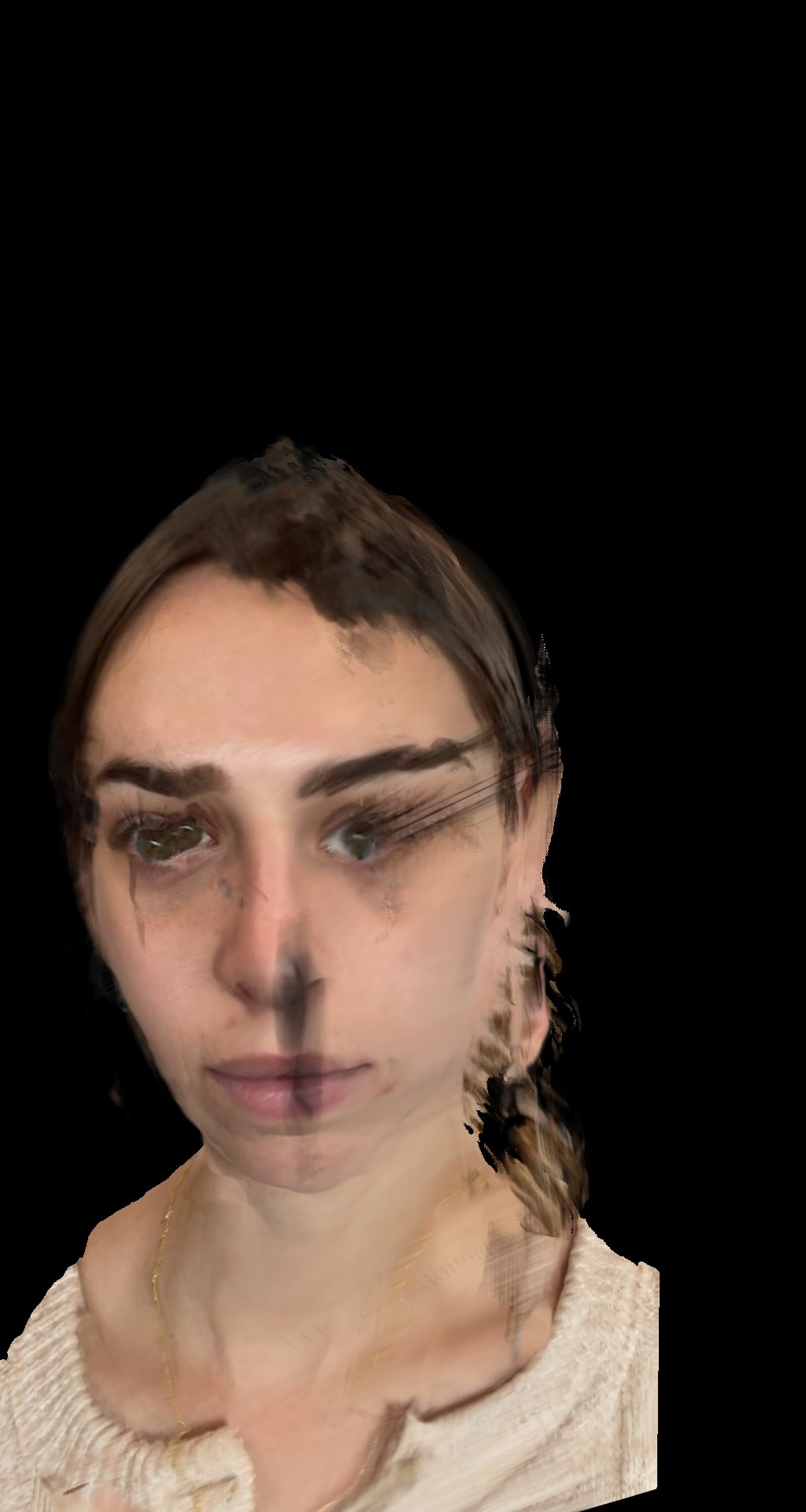} \\

            \scriptsize{\rotatebox{90}{\phantom{+++++}KeypointNeRF}}  &
            \includegraphics[width=\sz\linewidth, trim={0 150 0 300},clip]{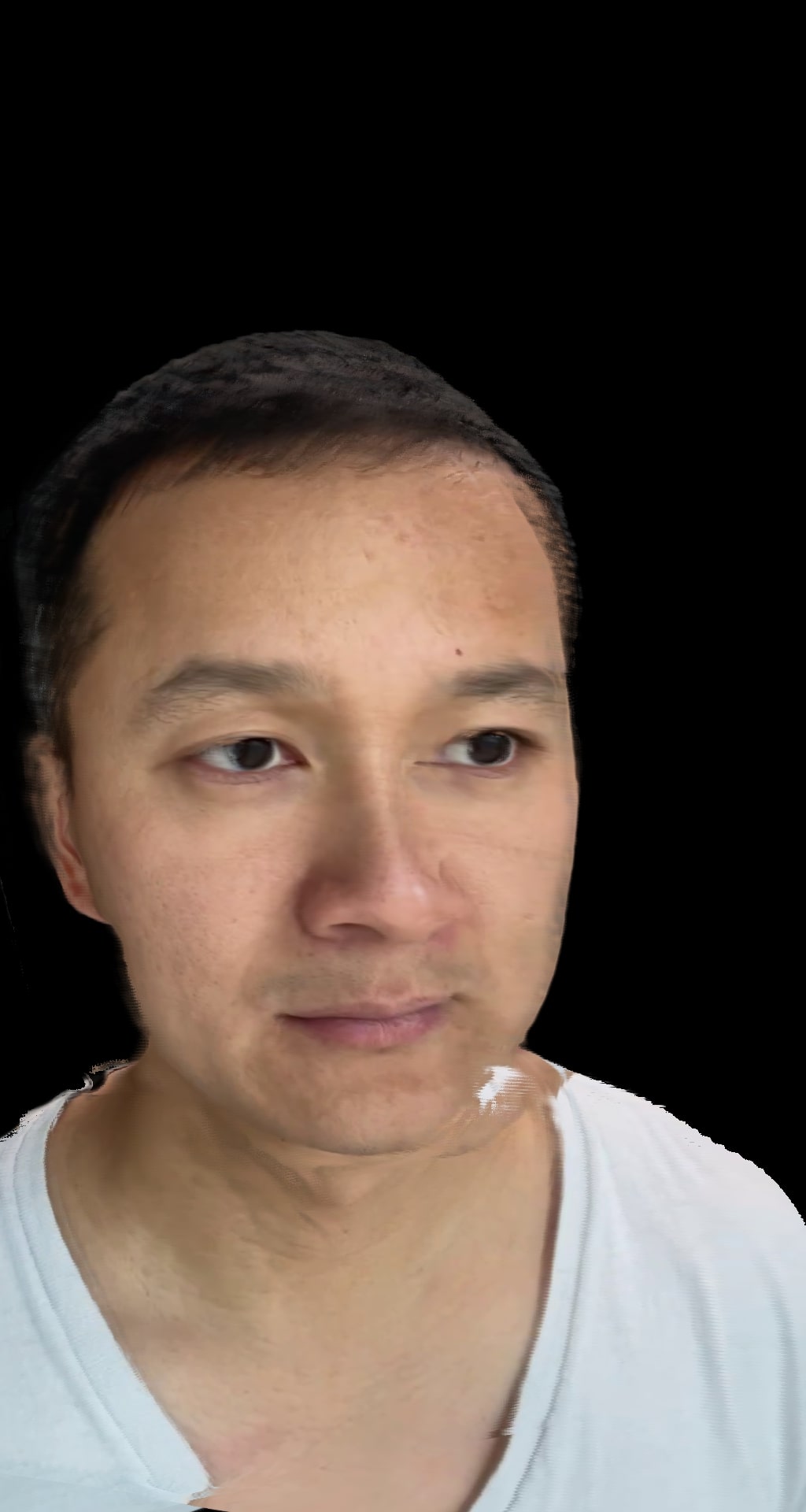} &
            \includegraphics[width=\sz\linewidth, trim={0 150 0 300},clip]{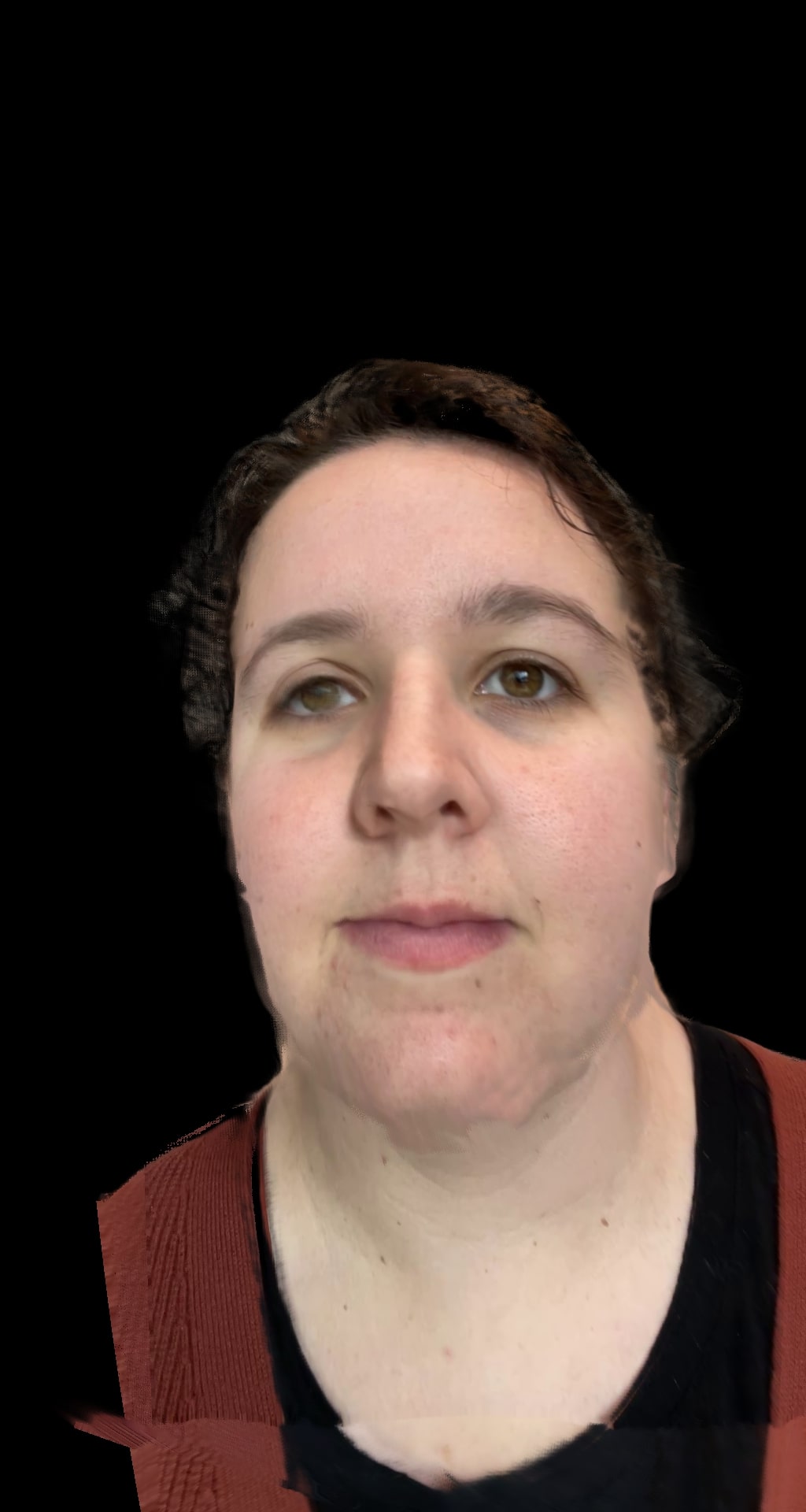} &
            \includegraphics[width=\sz\linewidth, trim={0 150 0 300},clip]{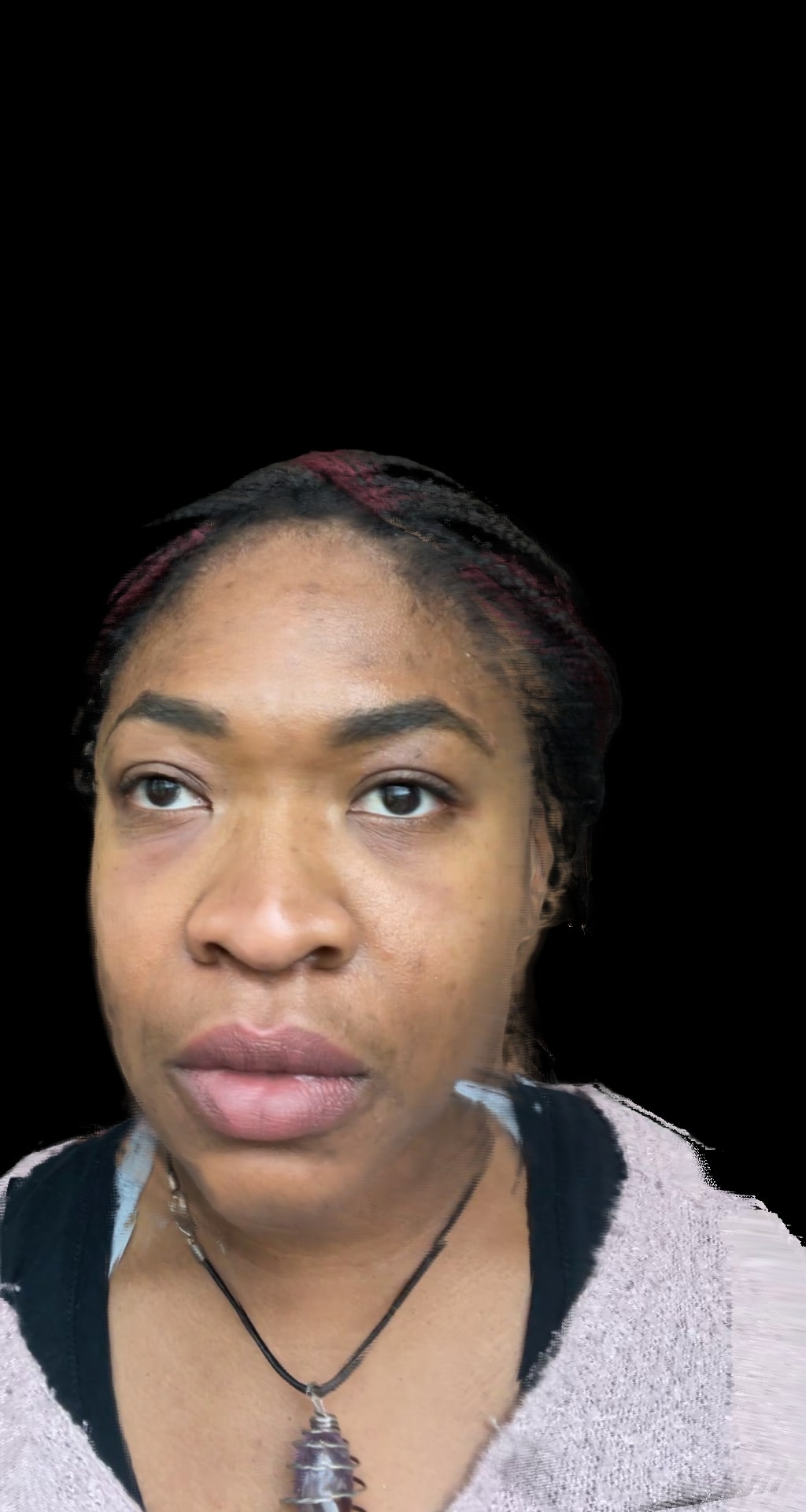} &
            \includegraphics[width=\sz\linewidth, trim={0 150 0 300},clip]{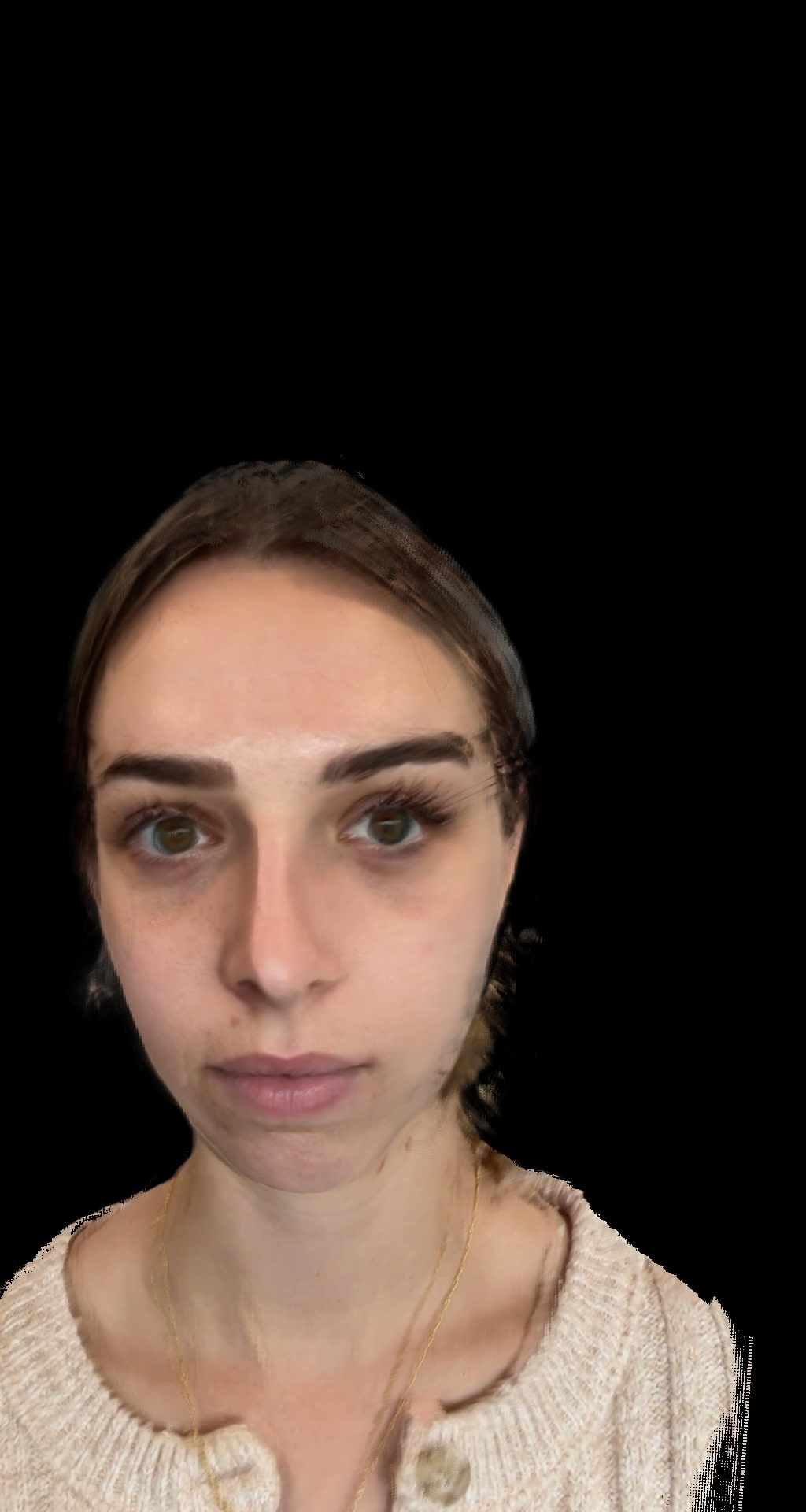} \\
            
            \scriptsize{\rotatebox{90}{\phantom{+++++++++}GT}}  &
            \includegraphics[width=\sz\linewidth, trim={0 150 0 300},clip]{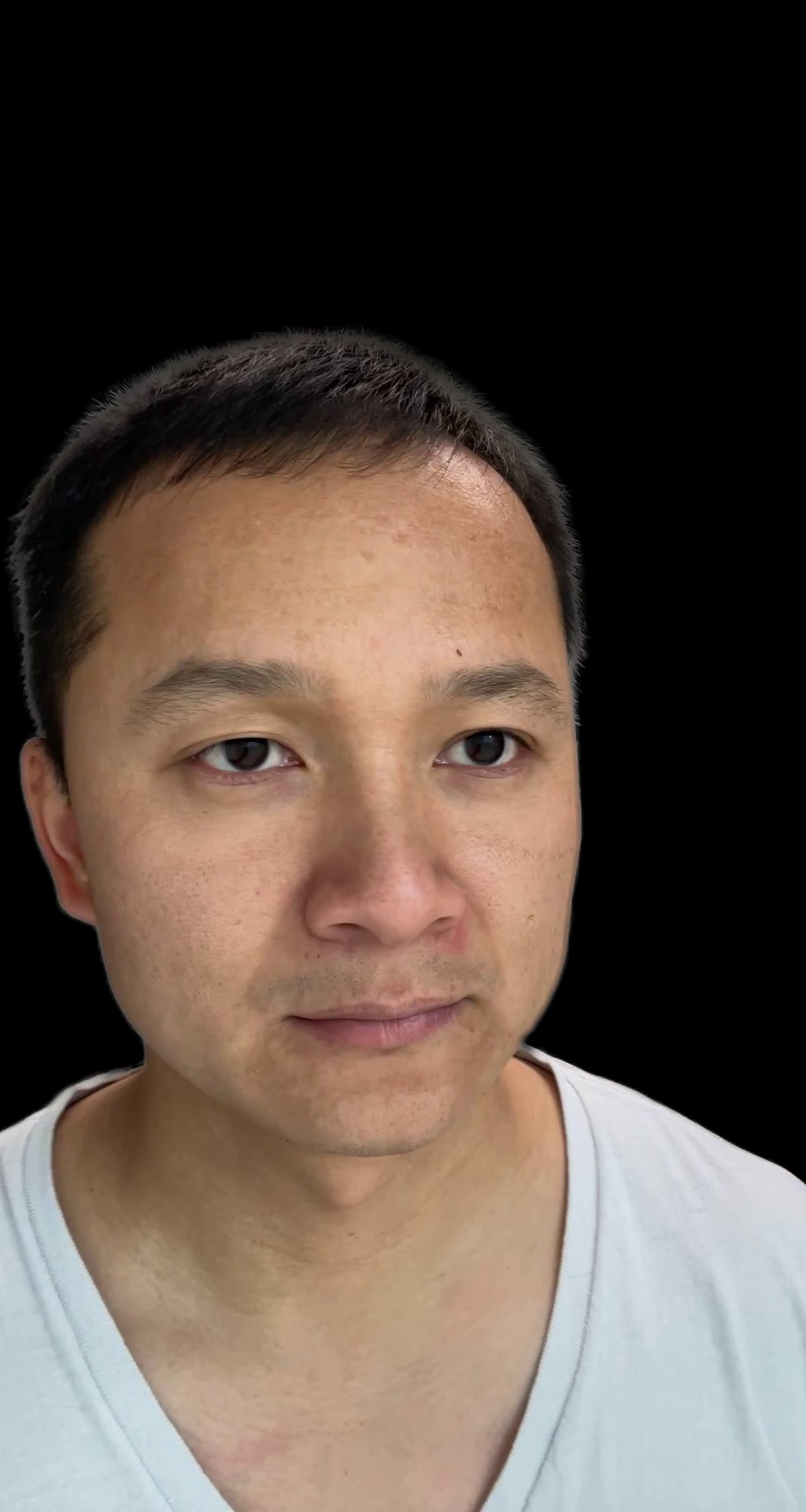} &
            \includegraphics[width=\sz\linewidth, trim={0 150 0 300},clip]{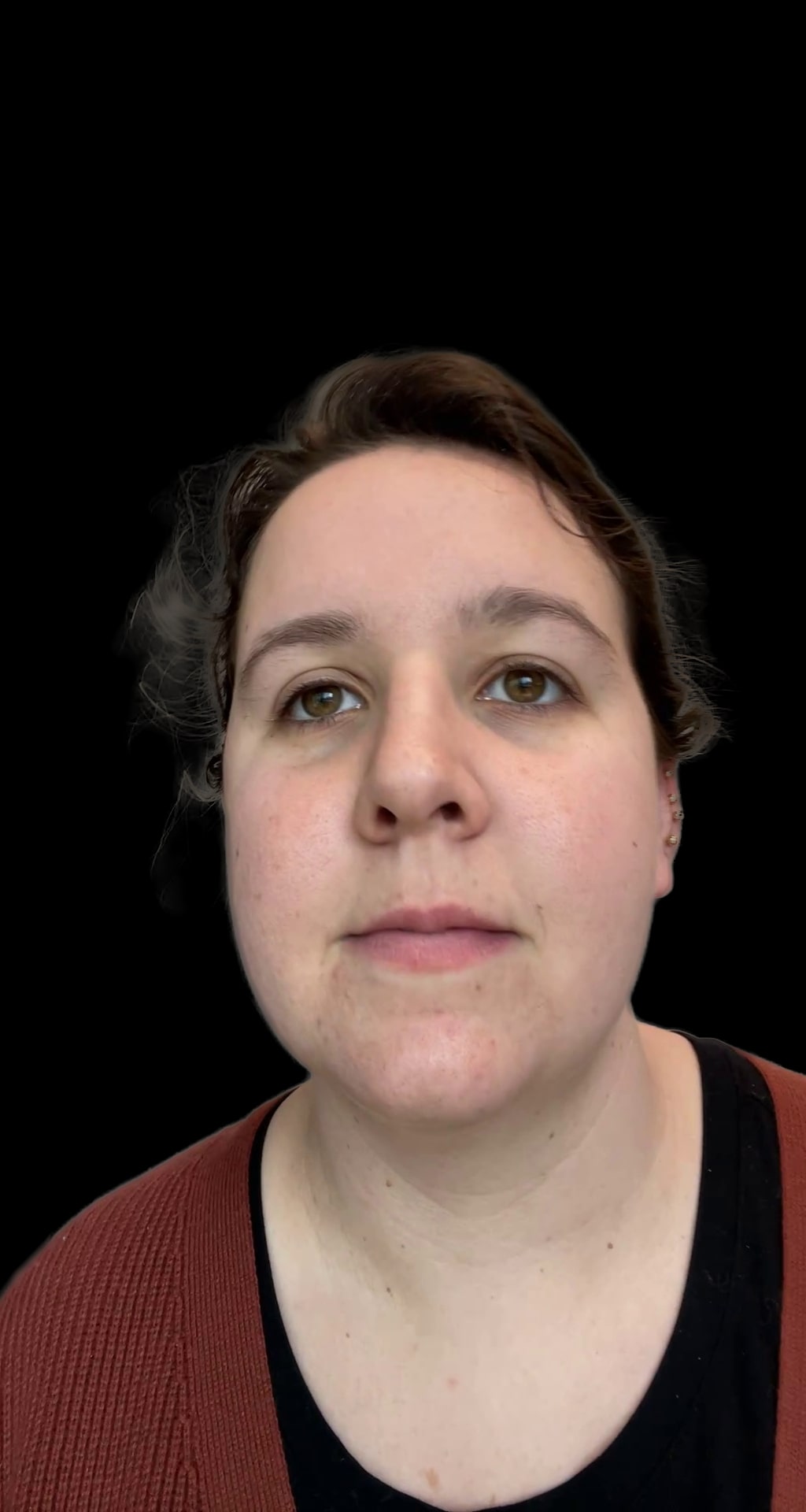} &
            \includegraphics[width=\sz\linewidth, trim={0 150 0 300},clip]{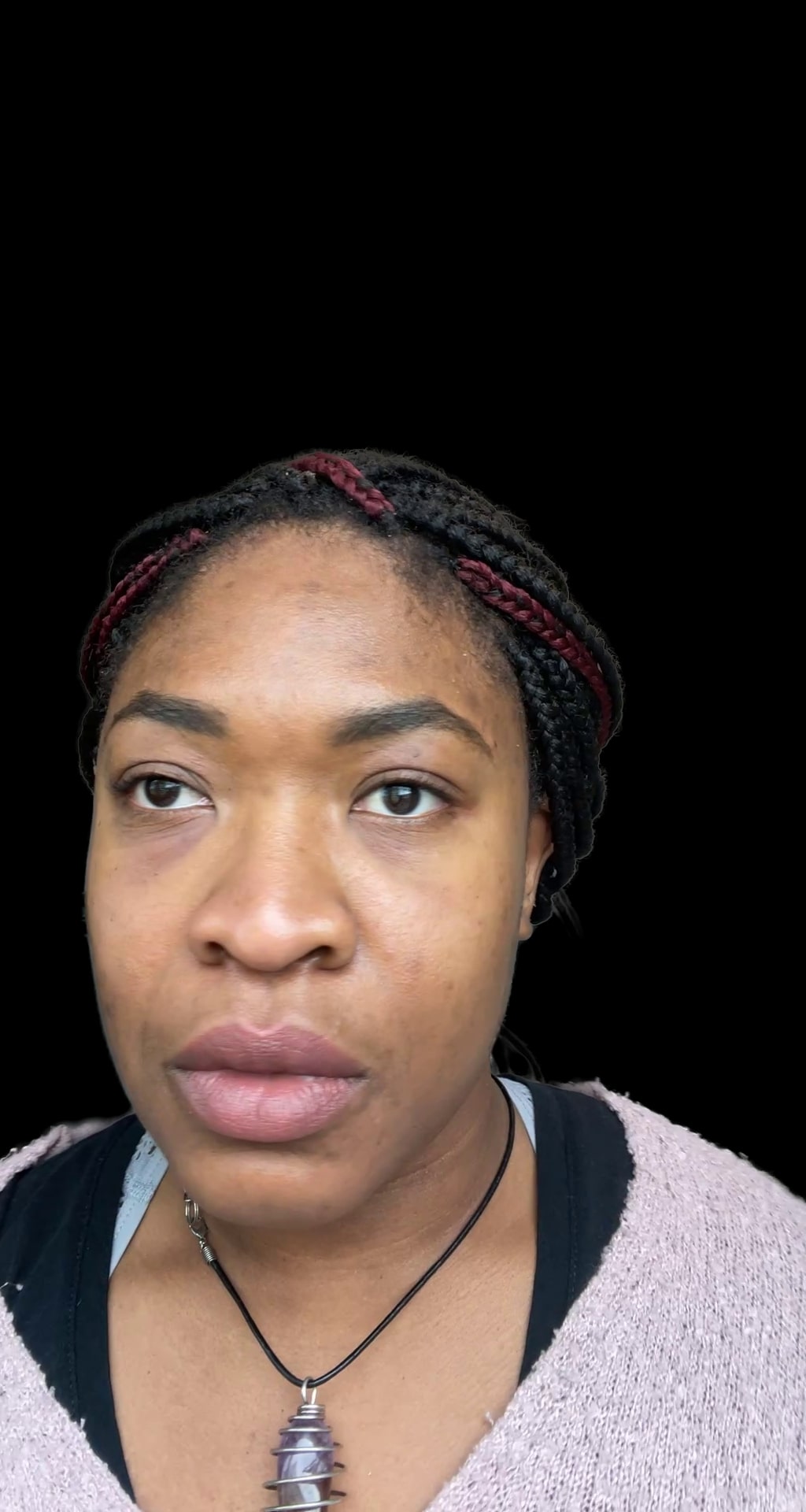} &
            \includegraphics[width=\sz\linewidth, trim={0 150 0 300},clip]{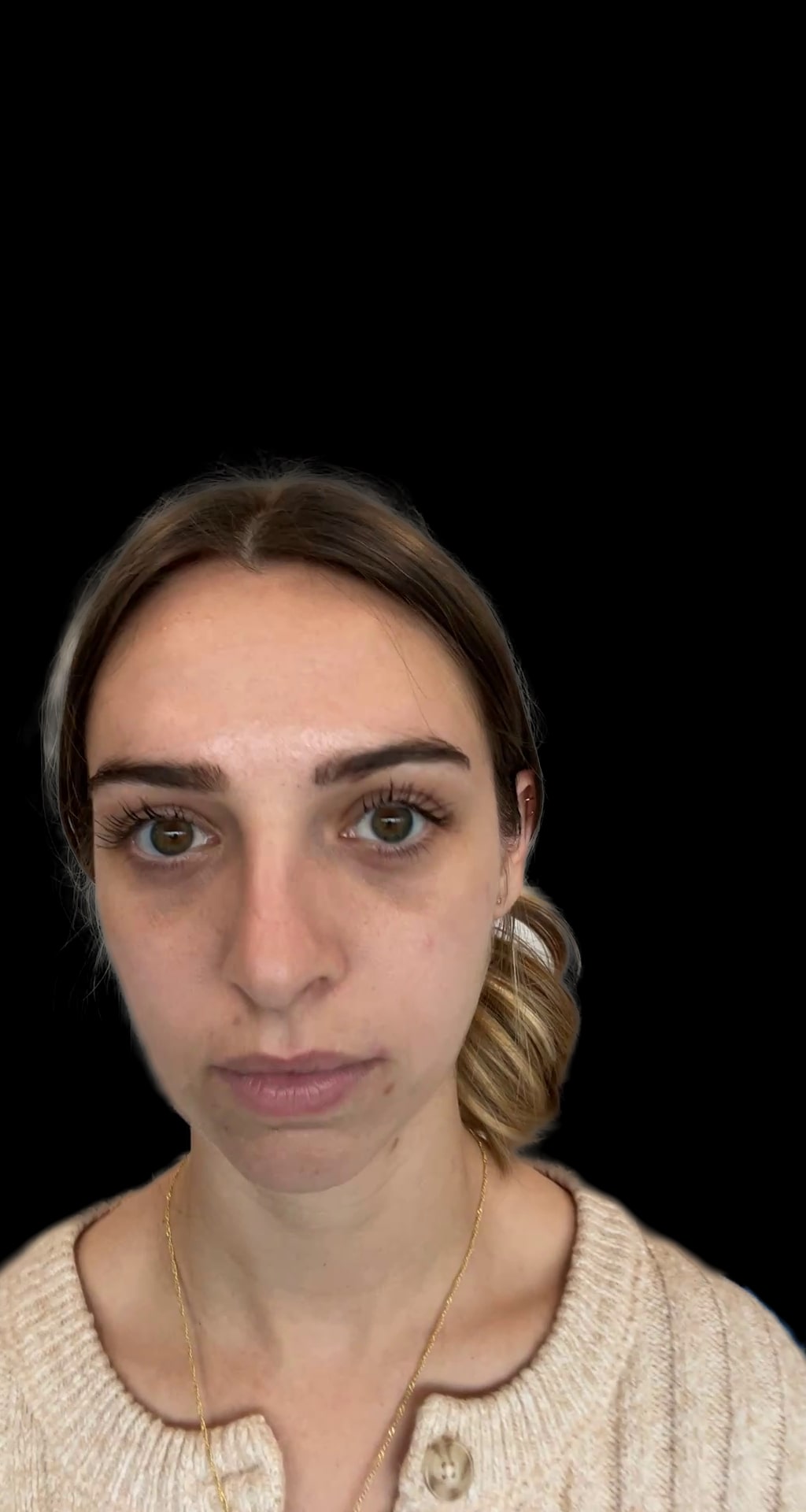} \\
    \end{tabular}
    \caption{\textbf{In-the-wild Captures.} Our approach produces high-quality reconstructions from three iPhone images, while all baselines show significant artifacts.}
    \label{fig:iphone_results}
\end{figure*}
\myparagraph{Setup.} To tackle the problem of reconstructing humans in the wild, we acquire a small dataset of subjects by taking several photos with an iPhone and estimate camera parameters.
We directly use intrinsic from manufacturing information of iPhone and extrinsic is computed by multi-view RGB-D fitting as in \cite{Chen2022authentic}. 
We evaluate the reconstruction methods trained on the studio captured data in Sec.~\ref{subsec:exp_facedome} without any retraining.

As input, all methods take three $1920 \times 1024$-resolution images of a person and predict a radiance field that is then rendered from novel views. 
In \figurename~\ref{fig:iphone_results}, we display rendered novel views of IBRNet~\cite{wang2021ibrnet}, our method without any spatial encoding, and our method with the proposed spatial encoding.
The baseline methods produce significantly worse results with lots of blur and cloudy artifacts, whereas 
\begin{wraptable}{r}{0.4\linewidth}
    \scriptsize
    \setlength{\tabcolsep}{4.2pt}

    \caption{\textbf{In-the-wild Captures.} 
    }
    \label{tab:iphone_results}
    \begin{tabular}{@{}lccc@{}}
        \toprule
                                                            & SSIM$\uparrow$ & PSNR$\uparrow$\\
        \midrule
        IBRNet~\cite{wang2021ibrnet}                        &   81.74   &   18.45 \\
        Ours (no keypoints)	                                &   79.50   &   19.79 \\
        KeypointNeRF	                                    &   \textbf{86.73}   &   \textbf{25.29} \\
        \bottomrule\\
    \end{tabular}
\end{wraptable}
our method can reliably reconstruct the human heads. 
This improvement is quantitatively supported by computed SSIM and PSNR on novel held-out views of the visualized four subjects (Tab.~\ref{tab:iphone_results}). 
This experiment demonstrates that our relative spatial encoding is the crucial component for cross-dataset generalization.  
Please see the supplementary material for more visualizations.

\subsection{Reconstruction of Human Bodies}
Additionally, we demonstrate that our method is suitable for reconstructing full volumetric human bodies without relying on template fitting of parametric human bodies \cite{loper2015smpl}.
We use the public ZJU~\cite{peng2021neuralbody} dataset in order to follow the experimental setup used in \cite{kwon2021neuralhumanperformer}, so that we could closely compare our method’s ability to reconstruct human bodies to the current state-of-the-art method without changing any experimental variables.
We follow the standard training-test split of frames and use a total of seven subjects for training and three for validation. At inference time, all methods use three input views. 
We compare our method with the generalizable volumetric methods: pixelNeRF~\cite{yu2021pixelnerf}, PVA~\cite{raj2021pva}, the current state-of-the-art Neural Human Performer (NHP)~\cite{kwon2021neuralhumanperformer}, and our method without weighting the relative spatial encoding in Eq.~\ref{eq:sp_encoding}. 
We report results on unseen identities for 438 novel views in \tablename~\ref{tab_fig:res_human} and side-by-side qualitative comparisons with NHP in \figurename~\ref{fig:res_human}. 
The results demonstrate that weighting the spatial encoding benefits reconstruction of human bodies as well. 
Our method is on par with the significantly more complex NHP, which relies on the accurate registration of the SMPL body model~\cite{loper2015smpl} and temporal feature fusion, whereas ours only requires skeleton keypoints. 

\begin{table}[!ht]
    \caption{\textbf{Human Body Experiment.} Comparison of our method with the baseline methods pixelNeRF\cite{yu2021pixelnerf}, PVA~\cite{raj2021pva}, and Neural Human Performer (NHP)~\cite{kwon2021neuralhumanperformer}; ``no w.'' in the table means our method without weighting the relative spatial encoding (Eq~\ref{eq:sp_encoding}) }
    \begin{minipage}{.6\textwidth} %
            \newcommand{\sz}{0.19}  
            \scriptsize
            \setlength{\tabcolsep}{0.5pt}
            \begin{tabular}{ccccc}  
                    pixelNeRF & PVA & NHP & Ours & GT \\
                   \includegraphics[width=\sz\linewidth]{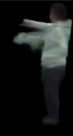} & 
                   \includegraphics[width=\sz\linewidth]{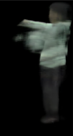} & 
                   \includegraphics[width=\sz\linewidth]{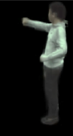} & 
                   \includegraphics[width=\sz\linewidth]{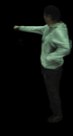} & 
                   \includegraphics[width=\sz\linewidth]{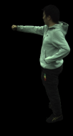} \\
            \end{tabular}

    \end{minipage} %
    \begin{minipage}{.3\textwidth} %
        \centering
        \scriptsize
        \setlength{\tabcolsep}{4.2pt}
        \begin{tabular}{@{}lcccc@{}}
            \toprule
                                      & PSNR$\uparrow$        & SSIM$\uparrow$        \\
            \midrule
            pixelNeRF                 & 23.17       & 86.93       \\
            PVA                       & 23.15       & 86.63       \\
            NHP                       & 24.75       & {\bf 90.58} \\
            Ours (no w.)              & 24.66       & 89.30       \\
            Ours                      & {\bf 25.03} & 89.69       \\
            \bottomrule\\
        \end{tabular}
        \label{tab_fig:res_human}
    \end{minipage}
\end{table}
\begin{figure*}[!ht]
    \scriptsize
    \setlength{\tabcolsep}{0.6mm} 
    \renewcommand{\arraystretch}{0}
    \newcommand{\sz}{0.95}  
    \centering
    \begin{tabular}{cc}  
           \rotatebox{90}{\phantom{++++}NHP~\cite{kwon2021neuralhumanperformer}} & \includegraphics[width=\sz\linewidth]{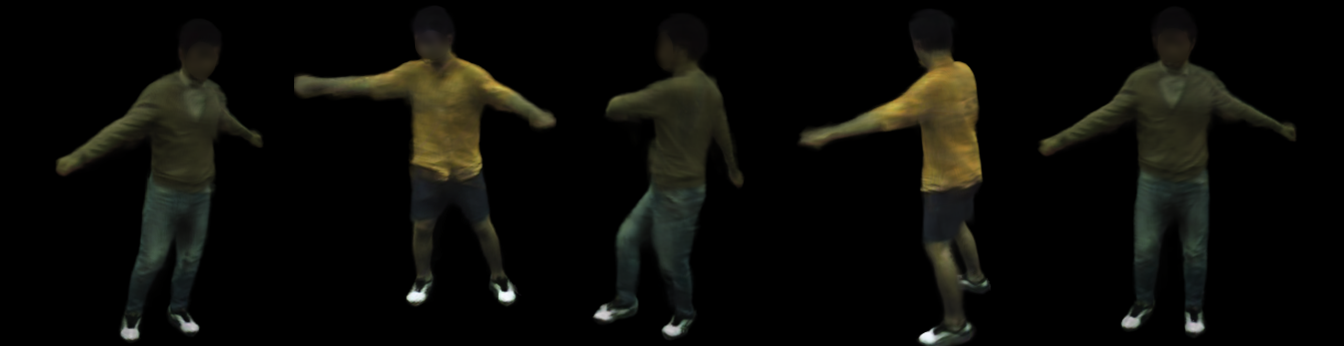} \\
           \rotatebox{90}{\phantom{+++}KeypointNeRF} & \includegraphics[width=\sz\linewidth]{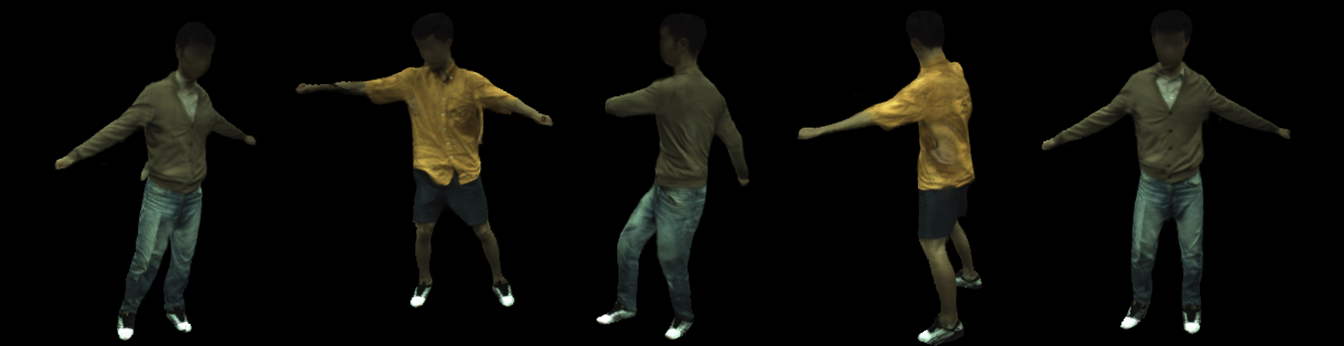} \\
           \rotatebox{90}{\phantom{+++++++}GT} & \includegraphics[width=\sz\linewidth]{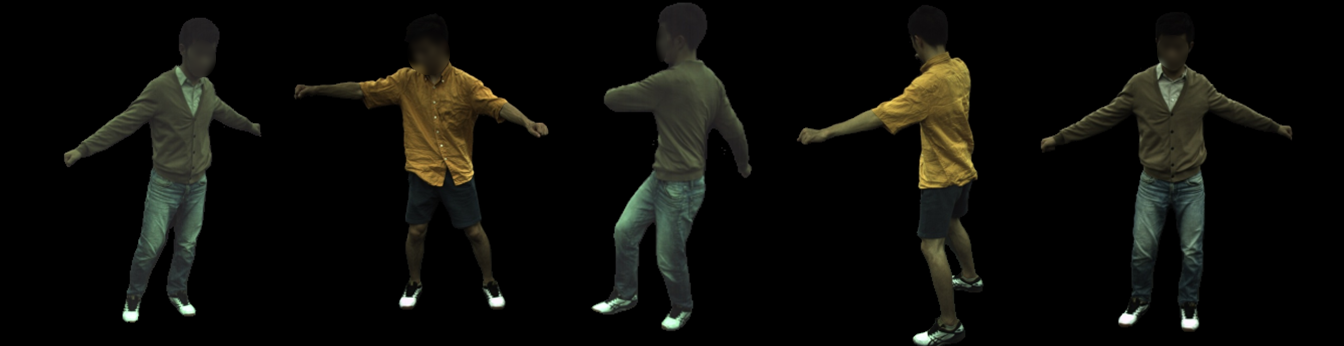} \\
    \end{tabular}
    \caption{\textbf{Human Body Experiment.} Comparison of NHP~\cite{kwon2021neuralhumanperformer} and our method on unseen identities from the ZJU-MoCap dataset~\cite{peng2021neuralbody}. Best viewed in electronic format.}
\label{fig:res_human}
\end{figure*}

\section{Conclusion}
We present a simple yet highly effective approach for generating high-fidelity
volumetric humans from as few as two input images.
The key to our approach is a novel spatial encoding based on relative information extracted from 3D keypoints.
%
Our approach outperforms state-of-the-art methods for head reconstruction and better generalizes to challenging out-of-domain inputs, such as selfies captured in the wild by an iPhone.
Since our approach does not require a parametric template mesh, it can be applied to the task of body reconstruction without modification, where it achieves performance comparable to more complicated prior work that has to rely on parametric human body models and temporal feature aggregation.
We believe that our local spatial encoding based on keypoints might also be useful for many other category-specific neural rendering applications. 

\myparagraph{Acknowledgments.}
We thank Chen Cao for the help with the in-the-wild iPhone capture. M. M. and S. T. acknowledge the SNF grant 200021 204840.

\bibliographystyle{splncs04}
\bibliography{egbib}
\clearpage
\appendix

\setcounter{page}{1}

\appendix
\setcounter{page}{1}
\setcounter{table}{0}
\setcounter{figure}{0}
\setcounter{equation}{0}
\renewcommand{\thetable}{\thesection.\arabic{table}}
\renewcommand{\thefigure}{\thesection.\arabic{figure}}
\renewcommand{\theequation}{\thesection.\arabic{equation}}
\title{KeypointNeRF:\\Generalizing Image-based Volumetric Avatars using Relative Spatial Encoding of Keypoints \newline --- Supplementary Material ---}

\titlerunning{KeypointNeRF}
\authorrunning{M. Mihajlovic et al.}

\author{}
\institute{}
\maketitle

\section{Overview}
In this document we provide 
additional implementation details (Sec.~\ref{sup:sec:impl_details}), 
information about the baseline methods (Sec.~\ref{sup:sec:baseline}), 
more qualitative and quantitative results (Sec.~\ref{sup:sec:results}), and 
reflect on the limitations of KeypointNeRF and future work (Sec.~\ref{sup:sec:limitations}). 

\section{Implementation Details} \label{sup:sec:impl_details}

\begin{figure*}[h!]
    \scriptsize
    \setlength{\tabcolsep}{0.5mm} 
    \newcommand{\sz}{0.24}  
    \begin{tabular}{ccccc}  
           \scriptsize{\rotatebox{90}{\phantom{+++}Inputs}} & 
           \includegraphics[width=\sz\linewidth]{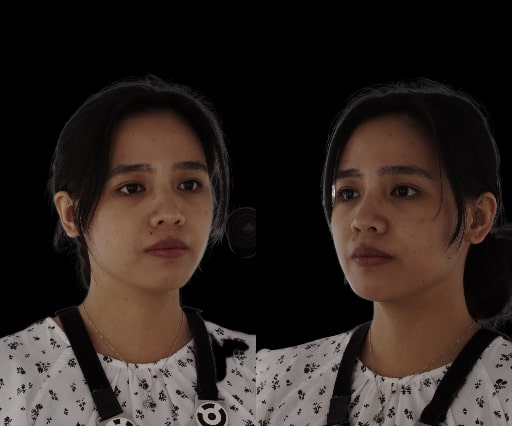} & 
           \includegraphics[width=\sz\linewidth]{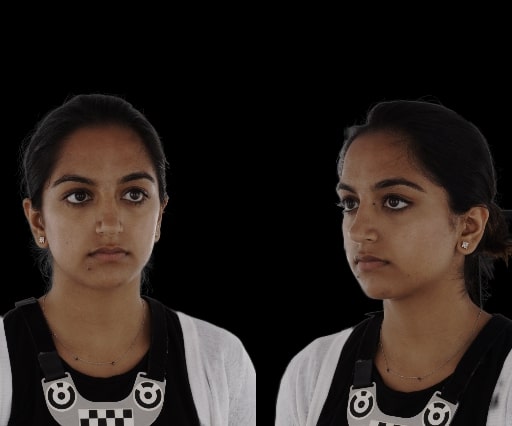} & 
           \includegraphics[width=\sz\linewidth]{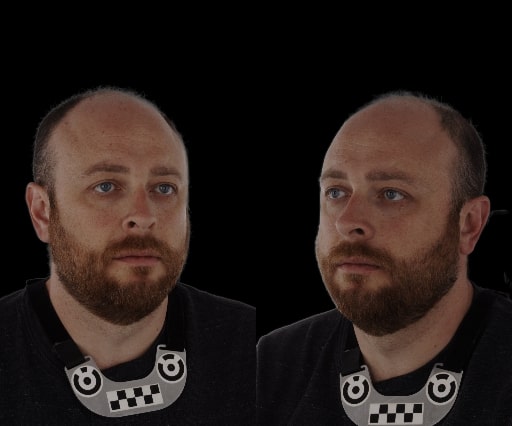} & 
           \includegraphics[width=\sz\linewidth]{figures/view2/ibrnet/332420187276318.front.jpg} \\

            \scriptsize{\rotatebox{90}{\phantom{+++}MVSNeRF~\cite{chen2021mvsnerf}}}  &
            \includegraphics[width=\sz\linewidth]{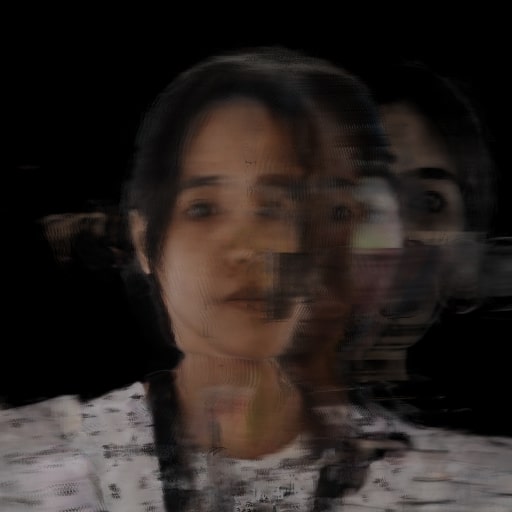} &
            \includegraphics[width=\sz\linewidth]{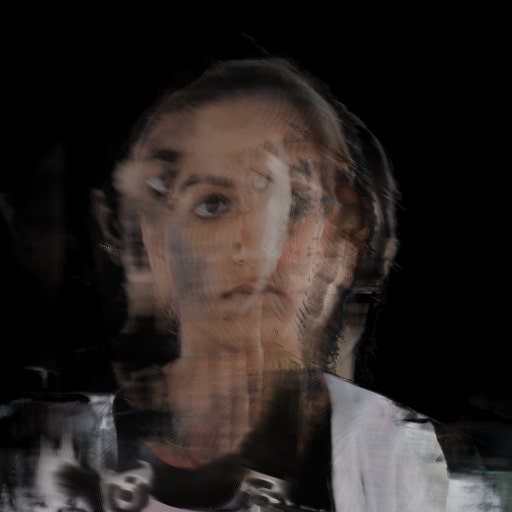} &
            \includegraphics[width=\sz\linewidth]{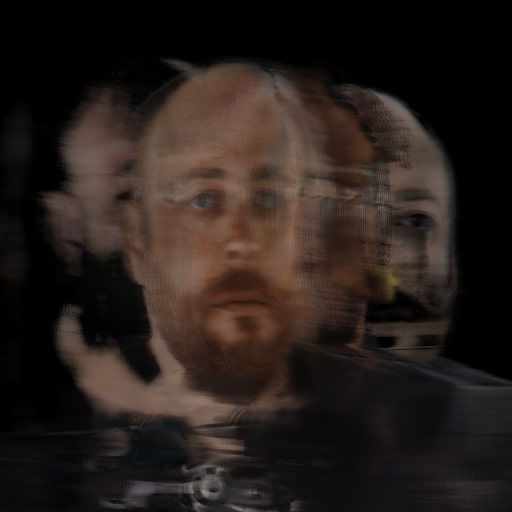} &
            \includegraphics[width=\sz\linewidth]{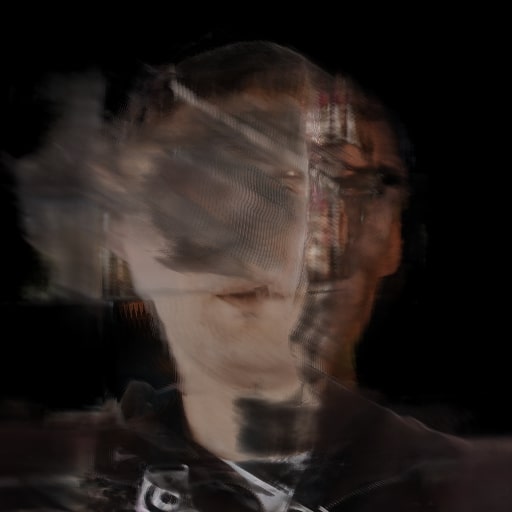} \\

            \scriptsize{\rotatebox{90}{\phantom{+++++}PVA~\cite{raj2021pva}}}  &
            \includegraphics[width=\sz\linewidth]{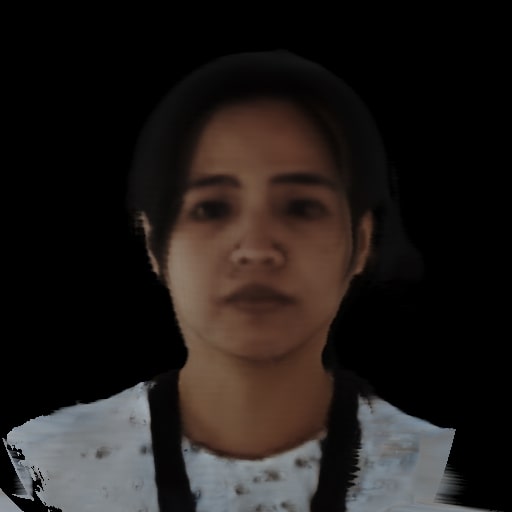} &
            \includegraphics[width=\sz\linewidth]{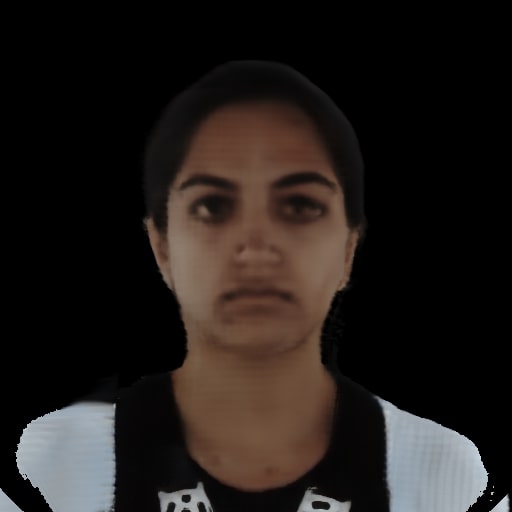} &
            \includegraphics[width=\sz\linewidth]{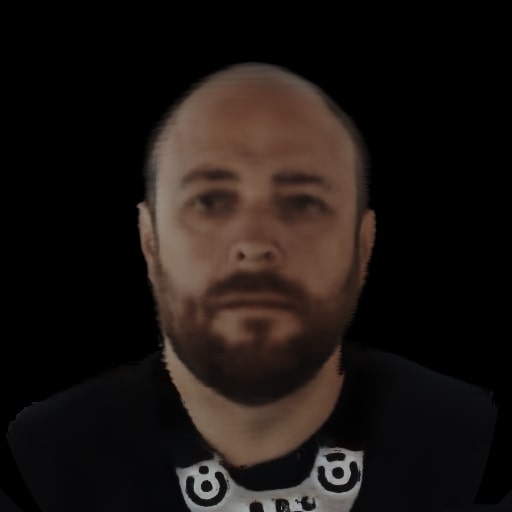} &
            \includegraphics[width=\sz\linewidth]{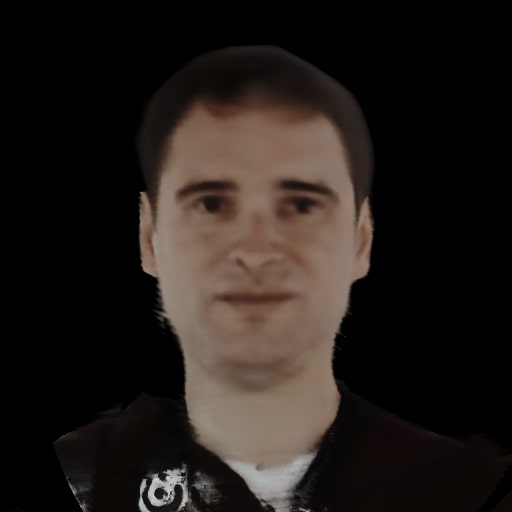} \\

            \scriptsize{\rotatebox{90}{\phantom{++++}IBRNet~\cite{wang2021ibrnet}}}  &
            \includegraphics[width=\sz\linewidth]{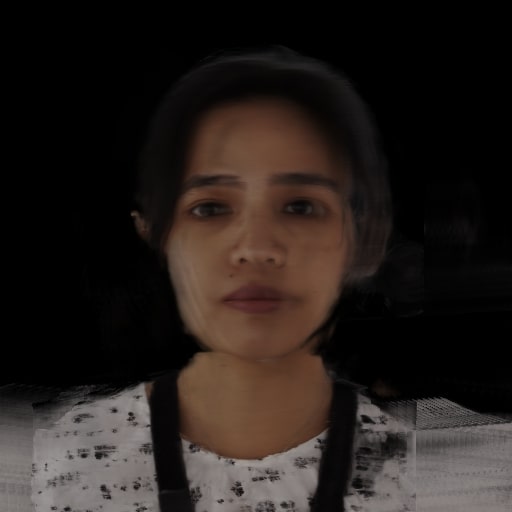} &
            \includegraphics[width=\sz\linewidth]{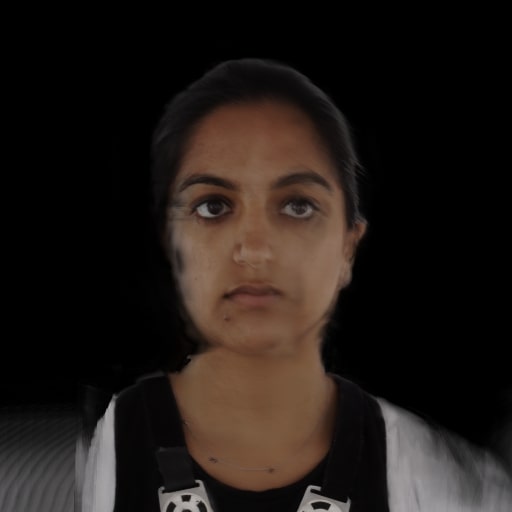} &
            \includegraphics[width=\sz\linewidth]{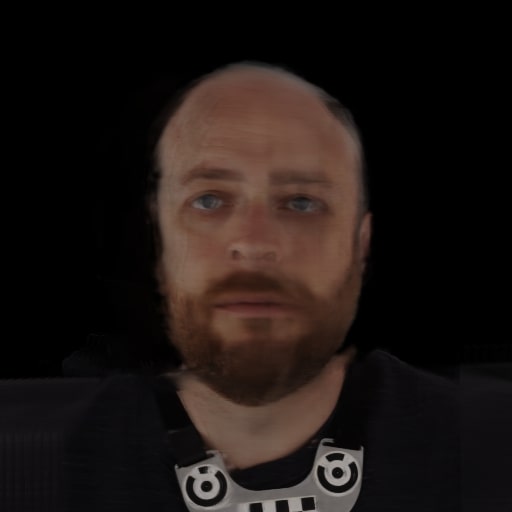} &
            \includegraphics[width=\sz\linewidth]{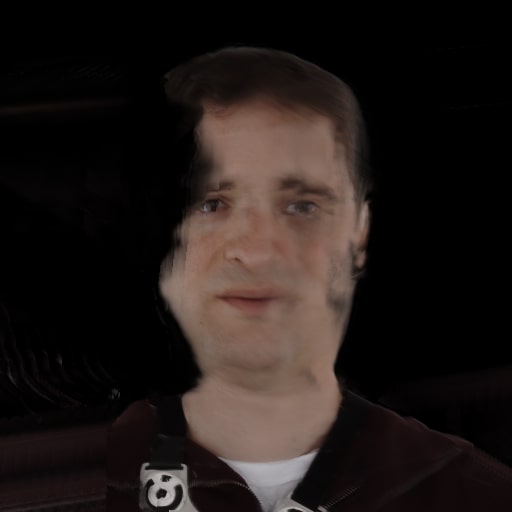} \\

            \scriptsize{\rotatebox{90}{\phantom{+++}KeypointNeRF}}  &
            \includegraphics[width=\sz\linewidth]{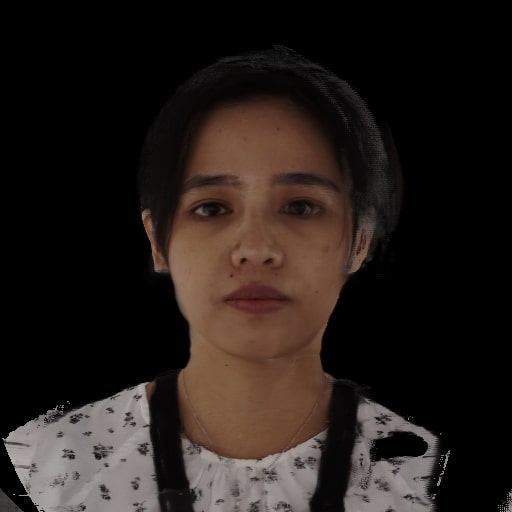} &
            \includegraphics[width=\sz\linewidth]{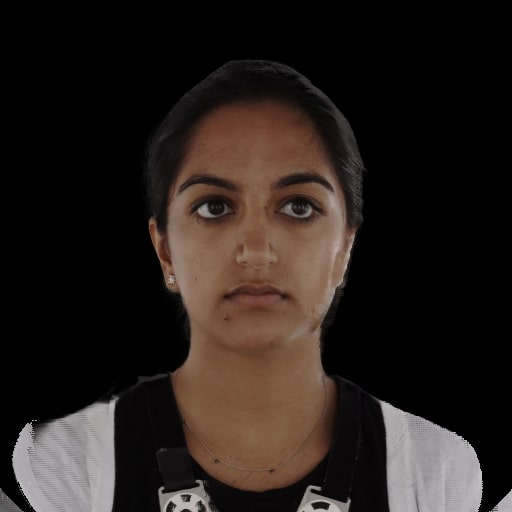} &
            \includegraphics[width=\sz\linewidth]{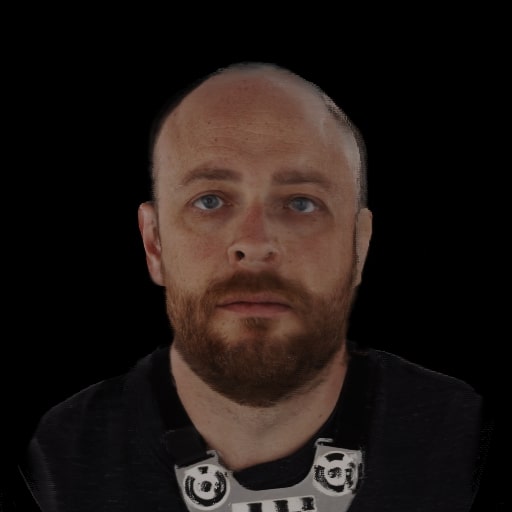} &
            \includegraphics[width=\sz\linewidth]{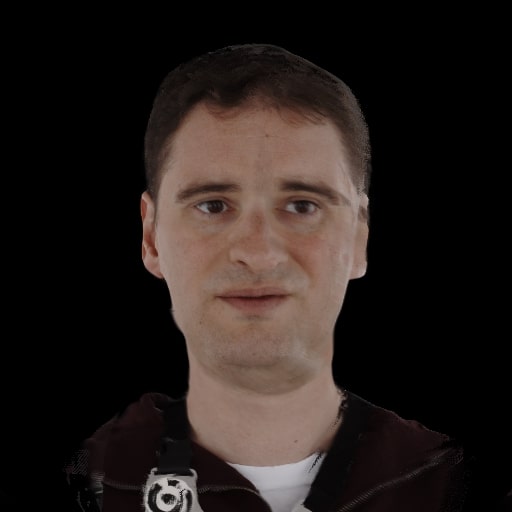} \\
            
    \end{tabular}

    \caption{\textbf{Studio Capture Results.} 
    Reconstruction results on held-out subjects from only two input views. 
    Our method produces much sharper results with fewer artifacts compared to prior work. 
    Best viewed in electronic format.
    }
    \label{fig:sup:results_2views}
\end{figure*}

\myparagraph{Image Encoders.} 
We employ a single HourGlass~\cite{hourglass} network to learn a geometric prior of humans and condition the density estimation network. 
The input image is normalized to $[-1, 1]$ range and processed by four convolutional blocks (256 filters) interleaved with group normalization.
We then employ an HourGlass block (down-sampling rate of four) with group normalization layers and refine the final output with four convolutional layers to produce the deep feature map $F^{gl}_n \in \mathbb{R}^{H/8 \times W/8 \times 64}$. 
Additionally, after the second convolutional block, we employ the transposed convolutional layer to produce the shallow high-resolution feature map $F^{gh}_n \in \mathbb{R}^{H/2 \times W/2 \times 8}$. 
As activation function we use ReLU for all layers. 
We implemented a second convolutional encoder that is independent of the density prediction branch to produce an alternative pathway for the appearance information $F^{a}_n \in \mathbb{R}^{H/4 \times W/4 \times 8}$ as in DoubleField~\cite{shao2021doublefield}. 
We follow the design of \cite{johnson2016perceptual} and implement this encoder as a 15-layer convolutional network with residual connections and ReLU activations. 

\myparagraph{Multi-view Feature Fusion.}
The feature fusion network is implemented as a four-layer MLP (128, 136, 120, and 64 neurons with Softplus activations) that aggregates features from multiple views. 
Its output is aggregated via mean-variance pooling \cite{wang2021ibrnet} to produce the geometry feature vector $G_X \in \mathbb{R}^{128}$.

\myparagraph{Density Fields.}
The geometry feature vector is decoded as density value $\sigma$ via a four-layer MLP (64 neurons with Softplus activations). 

\myparagraph{View-dependent Color Fields.}
To produce the final color prediction $c$ for a query point $X$, we implemented an additional MLP that predicts blending weights as an intermediate step which are used to blend the input pixel colors.
This network follows the design proposed in IBRNet to communicate information among multi-view features by using the mean-variance pooling operator. 
The per-view input feature vectors (described in Sec.~\ref{subsec:m_radiance_fields}) are first fused into a global feature vector via the mean-variance pooling operator. 
Then this feature is attached to the pixel-aligned feature vectors $\Phi(X|F^{a}_n)$ and propagated through a nine-layer MLP with residual connections and an exponential linear unit as activation to predict the blending weights (Eq.~\ref{eq:color_pred}).

\section{Baseline Methods} \label{sup:sec:baseline}
We used the publicly released code of MVSNeRF~\cite{chen2021mvsnerf} and IBRNet~\cite{wang2021ibrnet} with their default parameters. 
We re-implemented PVA~\cite{raj2021pva} since their code is not public and we directly used the public results of NHP~\cite{kwon2021neuralhumanperformer} for the experiments on the ZJU-MoCap dataset~\cite{peng2021neuralbody}.

\section{Additional Results} \label{sup:sec:results}

\myparagraph{Multi-view studio Capture Results.}
We further provide qualitative results for two more baseline methods (MVSNeRF~\cite{chen2021mvsnerf} and PVA~\cite{raj2021pva}) for the experimental setup described in Sec.~\ref{subsec:exp_facedome}.
The results in Fig.~\ref{fig:sup:results_2views} demonstrate that the best performing baseline (IBRNet) produces incomplete images with lots of blur and foggy artifacts.
PVA yields consistent, but overly smoothed renderings, while MVSNeRF does not work well for the widely spread-out input views.
For more qualitative results, we refer the reader to the supplementary video. 

\begin{figure*}[t!]
    \begin{center}
        \includegraphics[width=\textwidth]{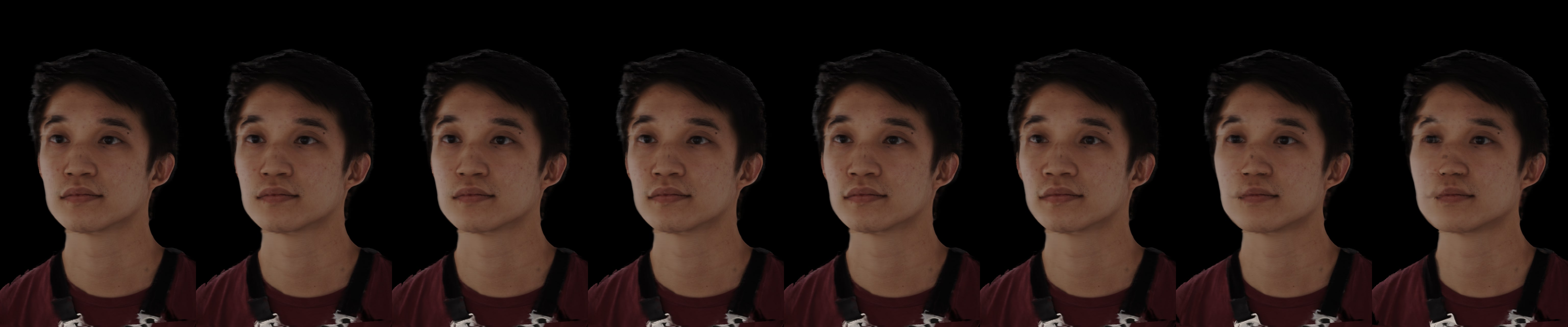}
    \end{center}
    \vspace{-0.6cm}
    \caption{
    \textbf{Keypoint perturbation} via different noise levels (from left to right: 1mm, 2mm, 3mm, 4mm, 5mm, 10mm, and 20mm). The rendered images tend to become blurry around the keypoints (e.g. eyes) for large noise levels ($> 10$mm).
    }
    \label{fig:sup:kpts_perturbation}
\end{figure*}

\myparagraph{Keypoint perturbation.}
To evaluate the sensitivity of our method on a less accurate estimation of keypoints, we perturb them with different Gaussian noise levels (ranging from 1 to 20mm) for unseen subjects from Sec.~\ref{subsec:exp_facedome} and observe that the rendered images (Fig.~\ref{fig:sup:kpts_perturbation}) occasionally tend to become blurry around the keypoints (e.g. eyes) for large noise levels ($> 10$mm).

\myparagraph{The impact of the iPhone calibration for the in-the-wild capture.}
We evaluate the robustness of KeypointNeRF to a nosier camera calibration by estimating the iPhone camera parameters without the depth term for the experimental setup presented in Sec.~\ref{subsec:iphone_results}. 
We observe (Tab.~\ref{tab:sup:iphone_results}) a negligible drop (PSNR/SSIM by -0.04/-0.5) in performance for our method, demonstrating the robustness of our method under noisy camera calibration. 
\begin{table*}
    \centering
    \setlength{\tabcolsep}{6.0pt}

    \caption{\textbf{In-the-wild Captures.} Quantitative comparison of IBRNet~\cite{wang2021ibrnet}, our method without any spatial encoding, and our method with the proposed keypoint encoding; visual results are provided in \figurename~\ref{fig:iphone_results} for the iPhone calibration with the depth term}
    \label{tab:sup:iphone_results}
    \begin{tabular}{@{}lcc|cc@{}}
        \toprule
                                                        & \multicolumn{2}{c}{RGB calibration}   & \multicolumn{2}{c}{RGB-D calibration}\\
                                                        & SSIM$\uparrow$ & PSNR$\uparrow$       & SSIM$\uparrow$   & PSNR$\uparrow$\\
        \midrule
        IBRNet~\cite{wang2021ibrnet}                    & 81.72 & 18.41                         &  81.74           &   18.45 \\
        Ours (no keypoints)	                            & 79.36 & 19.85                         &  79.50           &   19.79 \\
        KeypointNeRF	                                & \textbf{86.22} & \textbf{25.25}       &  \textbf{86.73}  &   \textbf{25.29} \\
        \bottomrule\\
    \end{tabular}

\end{table*}

\myparagraph{Convolutional feature encoders.}
We further measure the impact of the HourGlass feature extractor and compare it with the U-Net encoder that is used by the other baseline methods ~\cite{raj2021pva,wang2021ibrnet}. 
We follow the experimental setup from subsections~\ref{subsec:exp_facedome} and \ref{subsec:iphone_results} and report quantitative results in Tab.~\ref{tab:sup:encoders} and \ref{tab:sup:encoders_iphone} respectively. 
We observe that HourGlass encoder consistently improves the reconstruction quality. 

\begin{table*}
    \centering
    \setlength{\tabcolsep}{1.5pt}
    \caption{\textbf{Studio Capture Results.} HourGlass~\cite{hourglass} vs U-Net~\cite{wang2021ibrnet, raj2021pva} encoder for the experiment conducted in Sec.~\ref{subsec:exp_facedome}.}
    \label{tab:sup:encoders}
    \begin{tabular}{@{}clcc@{}}
        \toprule
        &                                               & SSIM$\uparrow$ & PSNR$\uparrow$ \\
        \midrule
        & PVA~\cite{raj2021pva}                                                 &   81.95   &   25.87 \\
        & IBRNet~\cite{wang2021ibrnet}                                          &   82.39   &   27.14 \\
        & KeypointNeRF (w. U-Net encoder~\cite{wang2021ibrnet, raj2021pva})          &	84.34   &	26.23 \\
        & KeypointNeRF (w. HourGlass encoder~\cite{hourglass})                  &	\textbf{85.19}   &	\textbf{27.64} \\
        \bottomrule\\
    \end{tabular}
\end{table*}
\begin{table*}
    \centering
    \setlength{\tabcolsep}{1.5pt}
    \caption{\textbf{In-the-wild Captures.} HourGlass~\cite{hourglass} vs U-Net~\cite{wang2021ibrnet, raj2021pva} encoder for the experiment conducted in Sec.~\ref{subsec:iphone_results}}
    \label{tab:sup:encoders_iphone}
    \begin{tabular}{@{}clcc@{}}
        \toprule
        &                                               & SSIM$\uparrow$ & PSNR$\uparrow$ \\
        \midrule
        & IBRNet~\cite{wang2021ibrnet}                                          &   81.72   &   18.41 \\
        & KeypointNeRF (w. U-Net encoder~\cite{wang2021ibrnet, raj2021pva})     &	84.20   &	\textbf{25.67} \\
        & KeypointNeRF (w. HourGlass encoder~\cite{hourglass})                  &	\textbf{86.22} & 25.25 \\
        \bottomrule\\
    \end{tabular}
\end{table*}

\section{Limitations and Future Work} \label{sup:sec:limitations}
\begin{figure*}
    \scriptsize
    \centering
    \setlength{\tabcolsep}{0.6mm} 
    \newcommand{\sz}{0.2}  
    \begin{tabular}{ccccc}  
           \scriptsize{\rotatebox{90}{\phantom{++}Inputs}} & 
           \multicolumn{2}{c}{\includegraphics[width=0.15\linewidth, trim={0 0 0 50},clip]{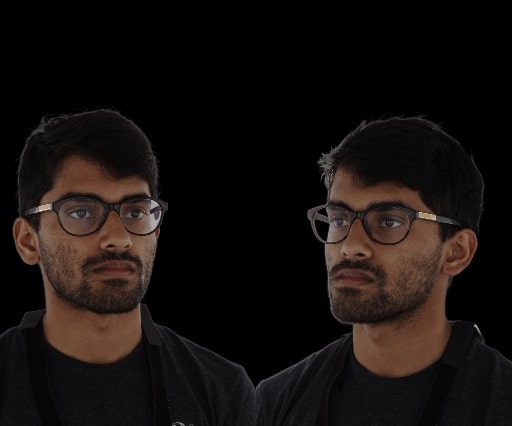}} &   
           \multicolumn{2}{c}{\includegraphics[width=0.2\linewidth, trim={0 10 0 10},clip]{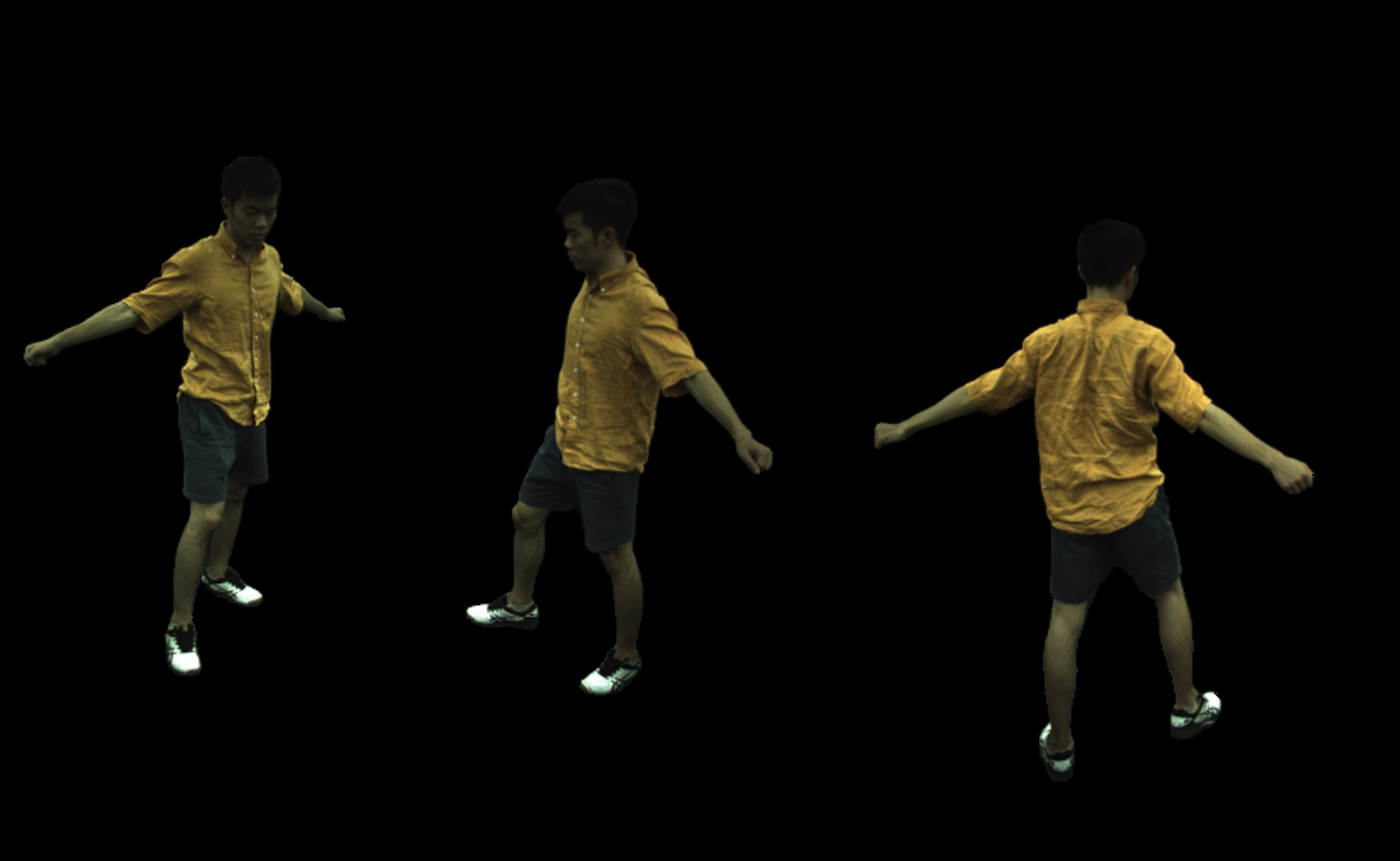}}  \\
           
            \scriptsize{\rotatebox{90}{\phantom{++}}} &
            \includegraphics[width=0.22\linewidth, trim={35 0 35 0},clip]{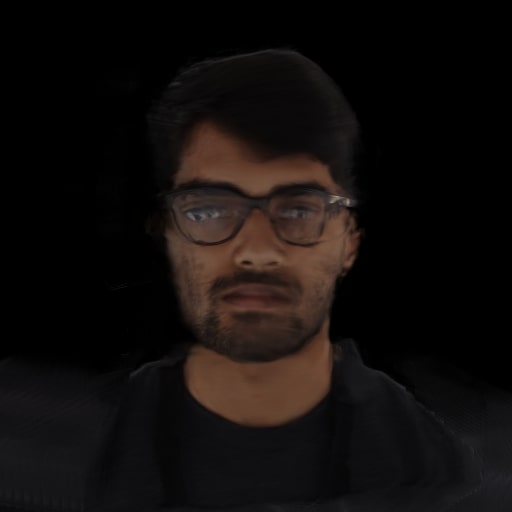} &
            \includegraphics[width=0.22\linewidth, trim={35 0 35 0},clip]{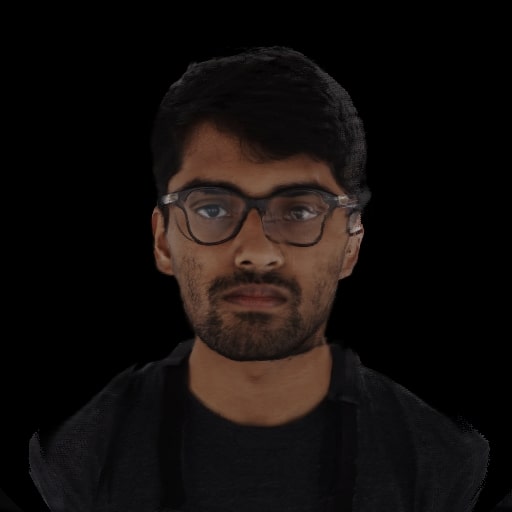} &
            \includegraphics[width=0.22\linewidth, trim={0 60 0 0},clip]{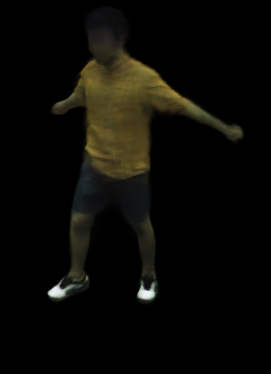} &
            \includegraphics[width=0.22\linewidth, trim={0 60 0 0},clip]{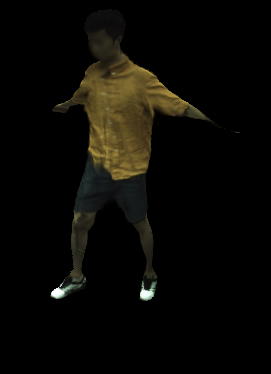} \\
            & IBRNet~\cite{wang2021ibrnet} & KeypointNeRF & NHP~\cite{kwon2021neuralhumanperformer} & KeypointNeRF \\

    \end{tabular}
    \caption{\textbf{Limitations.}
    Our method struggles to reconstruct the thin frames of the glasses (left) and has difficulties reconstructing human articulations that are outside of the training distribution. 
    }
    \label{fig:sup:limitations}
\end{figure*}

While our method offers an efficient way of reconstructing volumetric avatars from as few as two input images, it still has several difficulties.
The image-based rendering formulation of our method parametrizes the color prediction as blending of available pixels, 
which ensures good color generalization at inference time, however it makes the method sensitive to occlusions. 
The method itself has also difficulties reconstructing challenging thin geometries (e.g. glasses) and is less robust to highly articulated human motions (see Fig.~\ref{fig:sup:limitations}). 
As future work we consider addressing these challenges and additionally integrating learnable 3D lifting methods \cite{iskakov2019learnable, he2020epipolar} with the proposed relative spatial encoding for more optimal end-to-end network training. 

\end{document}